\documentclass[10pt,twocolumn,letterpaper]{article}

\usepackage[pagenumbers]{cvpr} 

\definecolor{cvprblue}{rgb}{0.21,0.49,0.74}
\usepackage[pagebackref,breaklinks,colorlinks,allcolors=cvprblue]{hyperref}

\usepackage{amsmath,amsfonts,bm}

\def\eqref#1{equation~\ref{#1}}

\def\1{\bm{1}}

\DeclareMathAlphabet{\mathsfit}{\encodingdefault}{\sfdefault}{m}{sl}
\SetMathAlphabet{\mathsfit}{bold}{\encodingdefault}{\sfdefault}{bx}{n}

\DeclareMathOperator*{\argmin}{arg\,min}

\usepackage{url}
\usepackage{graphicx} 
\usepackage{booktabs}
\usepackage{times}
\usepackage{epsfig}
\usepackage{graphicx}
\usepackage{amsmath}
\usepackage{amssymb}
\usepackage{pifont}
\usepackage{booktabs}
\usepackage{arydshln}
\usepackage{color}
\usepackage{colortbl}
\usepackage{subcaption}
\usepackage{multirow}
\usepackage{float}
\usepackage{rotfloat}
\usepackage{capt-of}
\usepackage{longtable}
\usepackage{diagbox}
\usepackage{makecell}
\usepackage{pgfplots}
\usepackage{xspace}
\usepackage{wrapfig}
\usepackage{enumitem}
\usepackage{fontawesome}
\usepackage{epigraph}
\usepackage{multirow}
\usepackage{multicol}
\usepackage{rotating}
\usepackage{amssymb}
\usepackage{pifont}
\usepackage{algpseudocode}
\definecolor{mycolor_blue}{HTML}{E7EFFA}
\definecolor{mycolor_green}{HTML}{E6F8E0}
\definecolor{mycolor_gray}{HTML}{ECECEC}
\definecolor{pearDark}{HTML}{2980B9}

\usepackage{fbox}
\definecolor{textcolor1}{rgb}{0.25,0.5,0.5}
\definecolor{textcolor2}{rgb}{0.7,0.25,0.25}
\definecolor{linkc}{rgb}{0, 0.44, 0.74}
\definecolor{eqc}{rgb}{1, 0, 0}
\definecolor{myy}{RGB}{126,95,0}
\definecolor{mygray}{gray}{.9}
\definecolor{bblue}{RGB}{30,80,120}
\definecolor{mygray1}{gray}{.7}
\definecolor{ggray}{RGB}{127,127,127}
\definecolor{mygreen}{RGB}{93,174,86}
\definecolor{citecolor}{HTML}{229954}

\usepackage{xcolor}

\usepackage[capitalize]{cleveref}
\crefname{section}{Section}{Secs.}
\Crefname{section}{Section}{Sections}
\Crefname{table}{Table}{Tables}
\crefname{table}{Tab.}{Tabs.}

\usepackage{indentfirst}
\usepackage{utfsym}

\def\paperID{****} 
\def\confName{CVPR}
\def\confYear{2025}

\title{TokenFlow: Unified Image Tokenizer for Multimodal Understanding and Generation}

\newcommand*\samethanks[1][\value{footnote}]{\footnotemark[#1]}
\author{Liao Qu\thanks{Equal contribution}, Huichao Zhang\samethanks, Yiheng Liu, Xu Wang\thanks{project lead}, Yi Jiang, Yiming Gao, Hu Ye, \\
Daniel K. Du, Zehuan Yuan, Xinglong Wu \\
ByteDance\\
\url{https://github.com/ByteVisionLab/TokenFlow}\\
}

\begin{document}
\maketitle

\begin{abstract}
We present \textbf{TokenFlow}, a novel unified image tokenizer that bridges the long-standing gap between multimodal understanding and generation. 
Prior research attempt to employ a single reconstruction-targeted Vector Quantization (VQ) encoder for unifying these two tasks. We  observe that understanding and generation require fundamentally different granularities of visual information. This leads to a critical trade-off, particularly compromising performance in multimodal understanding tasks.
TokenFlow addresses this challenge through an innovative dual-codebook architecture that decouples semantic and pixel-level feature learning while maintaining their alignment via a shared mapping mechanism. 
This design enables direct access to both high-level semantic representations crucial for understanding tasks and fine-grained visual features essential for generation through shared indices. 
Our extensive experiments demonstrate TokenFlow's superiority across multiple dimensions. Leveraging TokenFlow, we demonstrate for the first time that discrete visual input can surpass LLaVA-1.5 13B in understanding performance, achieving a 7.2\% average improvement. 
For image reconstruction, we achieve a strong FID score of 0.63 at 384×384 resolution.
Moreover, TokenFlow establishes state-of-the-art performance in autoregressive image generation with a GenEval score of 0.55 at 256×256 resolution, achieving comparable results to SDXL. 

\begin{figure}[htp]
\centering
\includegraphics[width=1.0\linewidth]{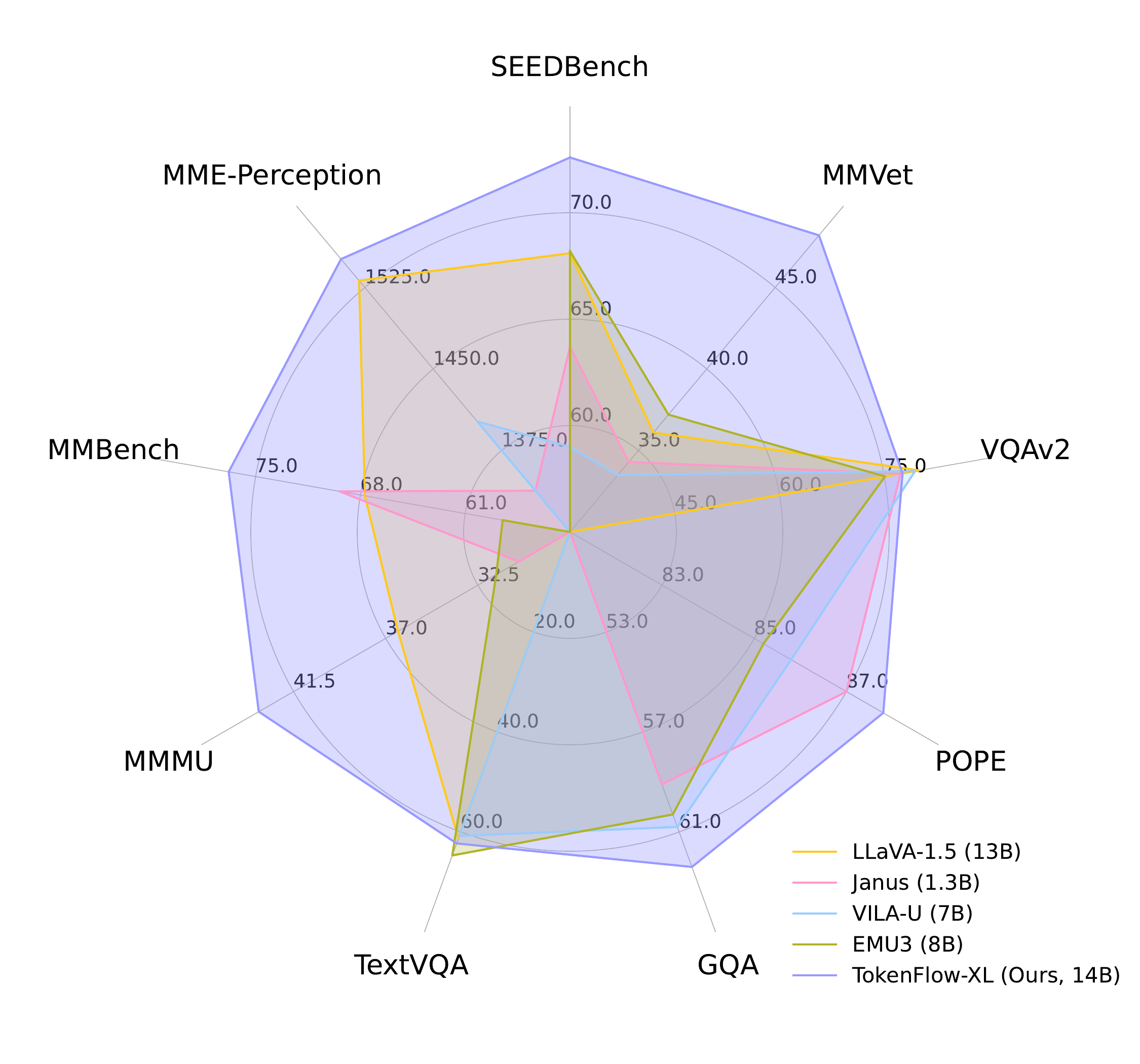}
\vspace{-1em}
\caption{Multimodal Understanding Results with TokenFlow. We demonstrate for the first time that \textit{discrete visual input} can surpass LLaVA-1.5 13B in understanding performance, achieving a 7.2\% average improvement.}
\label{fig:radar}
\end{figure}

\end{abstract}    
\section{Introduction}
\label{sec:intro}

\begin{figure*}[htp]
    \begin{center}
    \includegraphics[width=\linewidth]{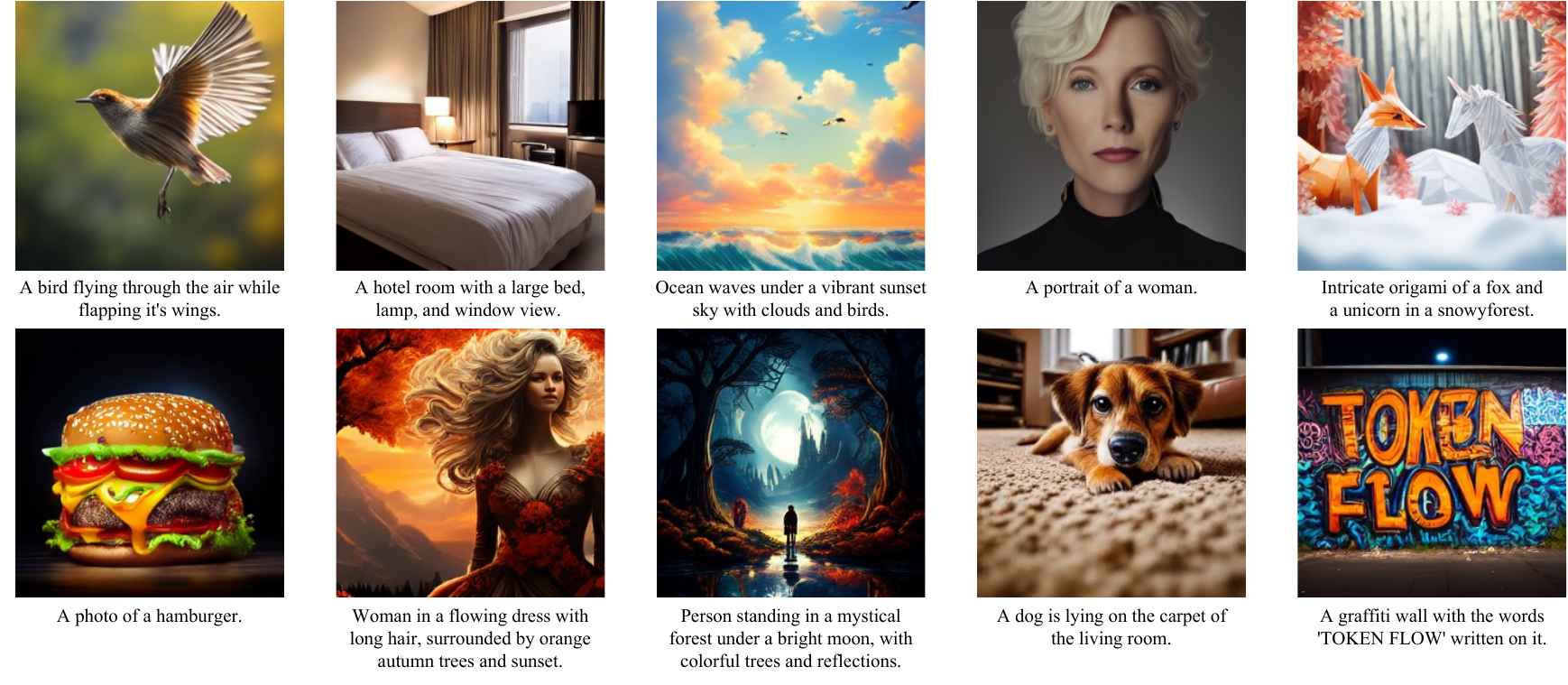}
    \caption{Visual Generation Results with TokenFlow. We present diverse 256×256 results across various styles, subjects, and scenarios.}
    \label{fig:gen_teasor}
    \end{center}
\end{figure*}

Large Language Models (LLMs) have revolutionized natural language processing through their unified autoregressive framework, demonstrating remarkable capabilities across diverse tasks \cite{gpt4, bai2023qwen}. 
However, in the multimodal domain of vision and language, a fundamental divide persists between perception and generation paradigms. Current approaches address them through distinct architectures: multimodal understanding models leverage vision encoders and projection layers to align visual representations with pretrained LLMs \cite{llava, tong2024cambrian}, while visual generation relies on either diffusion-based methods \cite{sd, dalle2} or discrete image tokens for autoregressive generation \cite{llamagen, parti, dalle1,var}. This divergence motivates the pursuit of unified approaches capable of both understanding and generation.

The advent of GPT-4o \cite{gpt4o} has greatly boosted interest in developing more generalist multimodal models. Early efforts to unify perception and generation capabilities \cite{emu1, minigemini} have primarily focused on equipping LLMs with the power of diffusion models. However, these approaches introduce substantial architectural complexity and computational overhead, highlighting the need for a more elegant unified solution. 
Recent efforts have explored one promising direction: using a single transformer architecture to unify visual and textual information within the next-token prediction framework \cite{wang2024emu3, team2024chameleon}. This approach relies on VQ encoders to convert visual inputs into discrete tokens that can be processed alongside text, offering a potentially simpler and more efficient framework. By treating both modalities as sequences of discrete tokens, this framework enables end-to-end training within a single architecture.

However, a fundamental challenge exists in such unified approaches. Multimodal understanding demands \textit{rich semantic representations} to support complex reasoning, while visual generation, on the other hand, requires \textit{precise encoding of spatial structure and textural details}. Current methods predominantly employ reconstruction-targeted VQ encoders \cite{zheng2022movq, vqgan}, which are primarily optimized for reconstruction fidelity. While this optimization makes them well-suited for generation tasks, it potentially limits their ability to capture the high-level semantic features crucial for understanding tasks. 
While Janus \cite{wu2024janus} attempts to address this conflict by employing separate encoders for understanding and generation tasks, this increases model complexity without fundamentally resolving the underlying representation disparity. 
These limitations underscore a critical gap in the field: the absence of a unified visual encoding mechanism that can effectively serve both perception and generation objectives.
This motivates our central research question: \textit{\textbf{Can one single image tokenizer derive representations suitable for both multimodal understanding and generation?}}

To address this challenge, we propose TokenFlow, a novel unified image tokenizer that bridges the gap between understanding and generation through a unique dual-flow design. The key insight is to decouple the learning of semantic and pixel-level features while maintaining their alignment through a shared index mapping. By mapping patches with both semantic and pixel-level similarities to identical indices, the quantized features can be directly applied to both autoregressive visual generation and multimodal understanding.
Unlike concurrent approach that constrains different feature levels within a single codebook \cite{vilau}, TokenFlow's dual-codebook design enables specialized learning while maintaining cross-level correlations through shared indices. This innovation allows simultaneous access to both semantic and pixel-level representations without compromising either aspect.
Specifically, TokenFlow adopts a dual-encoder architecture coupled with corresponding specialized codebooks. The semantic encoder, learned from a CLIP-style teacher, provides strong semantic priors, while the pixel encoder captures detailed visual information. The extracted features are then quantized by minimizing the weighted summation of semantic and pixel-level distances, creating a joint representation space.

Our framework exhibits remarkable scalability, maintaining exceptional codebook utilization (95\%+) even with large-scale codebooks of over 130K entries - substantially advancing beyond prior approaches \cite{vqgan} in both capacity and efficiency.
TokenFlow also achieves a strong FID score of 0.63 at 384×384 resolution.
For text-to-image synthesis, we establish a new state-of-the-art GenEval score of 0.55 at 256×256 resolution in the autoregressive paradigm while requiring significantly fewer sampling steps compared to existing methods like EMU3 \cite{wang2024emu3} and LlamaGen \cite{llamagen}.
On multimodal understanding benchmarks, TokenFlow achieves new state-of-the-art performance with minimal training overhead, surpassing LLaVA-1.5 13B by 7.2\% on average - for the first time discrete visual inputs can outperform this strong baseline.
These results validate TokenFlow's effectiveness as a unified visual tokenizer that bridges the long-standing gap between understanding and generation tasks.

\begin{figure*}[htp]
    \begin{center}
    \includegraphics[width=\linewidth]{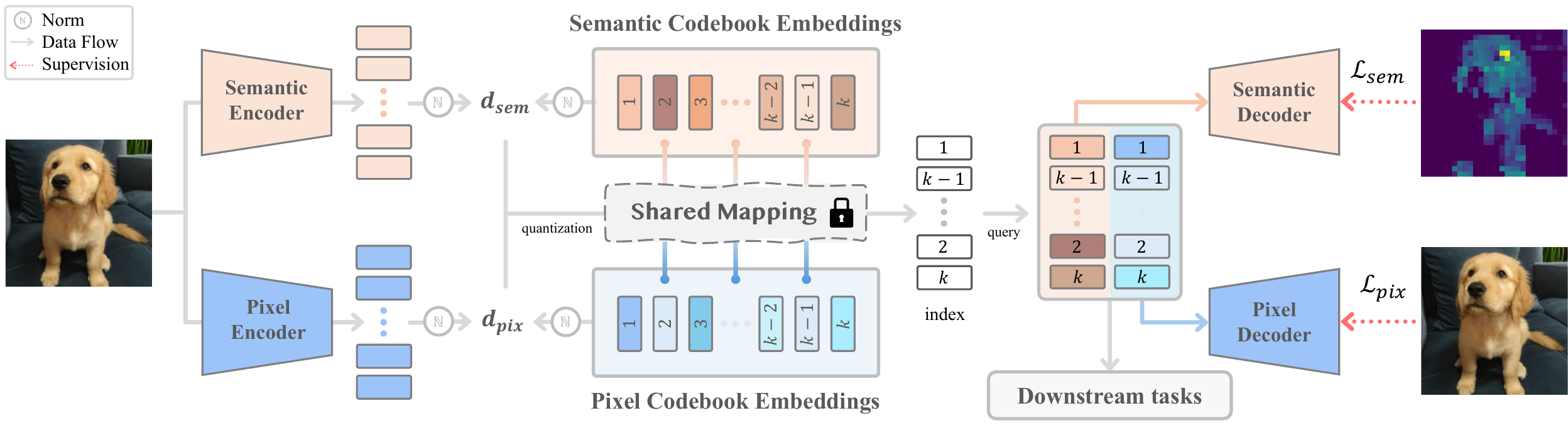}
    \caption{Overview of TokenFlow. We incorporate dual encoders and codebooks with a shared mapping, enabling the joint optimization of high-level semantics and low-level pixel details. For a given input image, distances $d_{\text{sem}}$ and $d_{\text{pix}}$ are calculated from the pixel-level and semantic-level codebooks, respectively, with the final codebook index and features determined by minimizing the weighted sum $d_{\text{sem}} + w_\text{dis} \cdot d_{\text{pix}}$. The resulting quantized features are independently decoded for both semantic alignment and image reconstruction training, and then concatenated to provide a unified representation for downstream tasks in understanding and generation.}
    \label{fig:pipeline}
    \end{center}
\end{figure*}

\section{Related Work}
\label{sec:rw}
\subsection{Tokenization for Visual Generation.}
Vector quantized (VQ) image tokenizers have played a crucial role in recent advancements in autoregressive image generation \citep{parti,var,llamagen,liu2024lumina,ma2024star}. \cite{vqvae} proposed the VQVAE, quantizing patch-level features using the nearest codebook entry, with the codebook learned with the encoder-decoder structure through reconstruction loss. VQVAE-2 \cite{vqvae2} advanced this framework through exponential moving average updates and a hierarchical multi-scale approach. VQGAN \citep{vqgan} further enhanced the architecture by incorporating adversarial and perceptual losses, yielding more precise and detailed representations.
Recent advances in VQ tokenizers have focused on three main directions: improving reconstruction fidelity and generation quality \citep{vitvqgan,rqvae,zheng2022movq}, enhancing codebook utilization \citep{vitvqgan,regvq,vqgan_lc}, and exploring novel architectures such as the multi-scale VQVAE \citep{var, li2024imagefolder} for next-scale prediction of images.
While these methods effectively preserve local details after quantization, they often struggle to capture semantic-level information, limiting their effectiveness in autoregressive multi-modal image understanding tasks. Our proposed TokenFlow addresses this limitation by introducing dual codebooks with shared mapping, achieving state-of-the-art performance in both autoregressive generation and multimodal understanding.

\subsection{Tokenization for Unified Multimodal Understanding and Generation}
Recent efforts have emerged to bridge the gap between multimodal understanding and generation \cite{li2024seed, team2024chameleon, showo, vilau, wang2024emu3, wu2024janus}.
Approaches like Chameleon \cite{team2024chameleon}, EMU3 \cite{wang2024emu3} and Show-o \cite{showo} employ VQ tokenizers \cite{vqgan, zheng2022movq, lfq} to encode images for both tasks. However, these methods typically require multimodal training from scratch and often suffer performance degradation in visual perception tasks due to limited semantic representation in their tokenized features.
SEED-LLaMA \cite{li2024seed} introduced a novel VQ tokenizer incorporating high-level semantics for understanding and utilize SD \cite{sd} as generation decoder. Janus \cite{wu2024janus} attempted to address the modality gap by employing separate tokenizers for understanding \cite{siglip} and generation \cite{llamagen}, though this leads to increased model complexity without fundamentally resolving the underlying challenge.
Concurrent work \cite{vilau} proposed a unified vision tower aligning discrete visual features with text during pre-training. However, their approach constrains low-level and high-level representations within a single flow, limiting the upper bound of downstream performance. 
In contrast, our work posits that the key to unifying understanding and generation lies in learning a universal mapping. By defining dual codebooks with shared mapping, TokenFlow enables flexible combinations of low and high-level features, resulting in superior performance across all downstream tasks.
\section{Method}
\label{3.method}

\subsection{Motivation}
\label{3.1}

\begin{table}[htp]
\small
\begin{center}
\caption{Comparison of various visual encoders on multimodal understanding \citep{textvqa,li2024seed,fu2023mme} within the LLaVA-1.5 framework. VQKD is distilled from CLIP ViT-B/14. "Sem." refers to semantic encoders that learn semantic-level representations, while "Pix." indicates pixel-level tokenizers that focus on low-level visual features.}
\label{wrap-tab:llava-vq-model}
\resizebox{\linewidth}{!}{
\begin{tabular}{clcccc}
    \toprule  
    \# Exp. & Visual Encoder & Type & MME-P $\uparrow$ & SEEDB $\uparrow$ & TQA $\uparrow$ \\
    \midrule
    \multicolumn{6}{l}{\textbf{\textit{Continuous:}}} \\
    1 & CLIP ViT-B/14~\citep{clip} & Sem. & 1460.9 & 64.1 & 53.4 \\
    \midrule
    \multicolumn{6}{l}{\textbf{\textit{Discrete:}}} \\
    2 & VQGAN~\citep{vqgan} & Pix. & 756.1 & 38.2 & 46.8 \\
    3 & VQGAN-LC~\citep{vqgan_lc} & Pix. & 744.8 & 38.2 & 45.7 \\
    4 & LFQ~\citep{lfq} & Pix. & 889.5 & 41.1 & 46.4 \\
    5 & VQKD~\citep{peng2022beit} & Sem. & 1252.4 & 57.8 & 48.2 \\
    \bottomrule
\end{tabular}
}
\end{center}
\end{table}

\begin{figure}[htp]
    \begin{center}
    \includegraphics[width=1.0\linewidth]{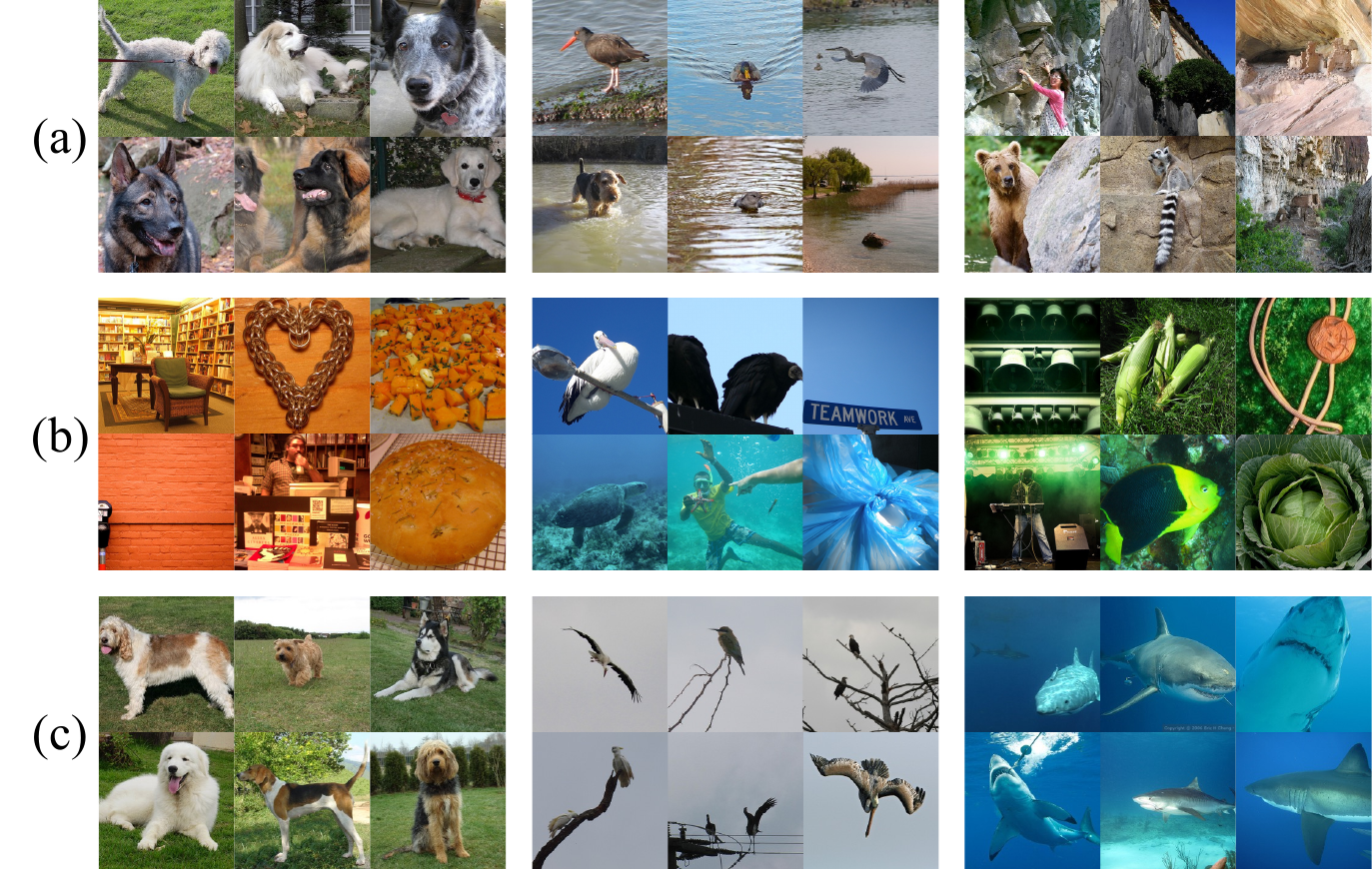}
    \caption{Visualization of images clustered by (a) VQKD \cite{peng2022beit}, (b) VQGAN \cite{vqgan}, and (c) Our TokenFlow. VQKD clusters exhibit semantic similarity, while VQGAN clusters exhibit low-level similarity (\textit{i.e.} color). Our TokenFlow can successfully combine both semantic and low-level similarity. Implementation details of image clustering can be found in \cref{appendix:1}.}
    \label{fig:diff-vqgan-beitv2}
    \end{center}
\end{figure}

Unifying multimodal understanding and generation into a cohesive next-token prediction paradigm requires a VQ tokenizer for extracting indices from input images.
While traditional VQ tokenizers \citep{vqvae,vqgan,vqgan_lc,lfq} excel at pixel-level image reconstruction, our investigation reveals a significant limitation in their image understanding capabilities. We conducted experiments utilizing these tokenizers as feature extractors within the LLaVA-1.5 \citep{llava} framework. As shown in Exp. 2-4 of \cref{wrap-tab:llava-vq-model}, the performance of these discrete tokenizers consistently lags behind that of the continuous tokenizer CLIP ViT-B/14 \citep{clip}. We posit that this performance gap stems from their pre-training objectives, which primarily optimize towards better low-level reconstruction quality. Consequently, the extracted features mainly encode low-level information, lacking the semantic-level understanding, which is crucial for complex visual reasoning.

Another straight forward solution for unified understanding and generation can be distill discrete tokens from pretrained CLIP \cite{clip,siglip,evaclip,chen2024vitamin}, and then equip it with image reconstruction capability. As demonstrated in Exp. 5, VQKD, distilled from CLIP ViT-B/14, substantially reduces the performance gap compared to other discrete tokenizers. 
We further conducted an experiment to reconstruct the original image from quantized features extracted by VQKD. The reconstructed images exhibited significant blurring and a evident loss of high-frequency details, as shown in \cref{fig:reconstruction-from-beitv2}. We attribute this outcome to the nature of VQKD's encoder, which maps semantically close patches into same codebook index. As visualized in \cref{fig:diff-vqgan-beitv2} (a), it tends to map images with same semantical meaning to the same codebook index, while VQGAN (\cref{fig:diff-vqgan-beitv2} (b)) tends to map visually similar images to the same codebook index, prioritizing low-level features over semantic content. 
Therefore, the reconstruction of fine-grained details from low-level dissimilar patches aggregated by VQKD becomes extremely challenging.

These observations highlight the necessity of developing a novel tokenization approach that can effectively handle high-level semantic understanding and low-level visual reconstruction tasks.

\subsection{Unified Image Tokenizer}
\label{3.2}
To bridge this gap, we propose TokenFlow (\cref{fig:pipeline}), a novel unified image tokenizer that enables joint representation learning at both semantic and pixel level.
We find the key to unifying understanding and generation lies in learning an universal mapping. If the tokenizer can map patches that are both high-level and low-level similar to the same codebook index, then the quantized features can be easily decoded and directly applied to both autoregressive visual generation tasks and multimodal understanding tasks.

\textbf{Encoder.}
Unlike previous approaches that utilize one single encoder to extract low-level image information, we propose a dual-encoder architecture comprising a semantic encoder $\mathcal{E}_\text{sem}$ and a pixel encoder $\mathcal{E}_\text{pix}$. This design enables the extraction of two distinct types of image features.
For the semantic encoder, we initialize it with a pre-trained text-aligned vision encoder (e.g., CLIP ViT-B/14). This initialization strategy facilitates better learning of high-level text-aligned embeddings in the semantic codebook, ultimately enhancing the model's multimodal understanding capabilities.
For brevity here, we omit the spatial indices of feature representations, where $\hat{z}_{\text {sem}} =\mathcal{E}_\text{sem}(x) \in \mathbb{R}^{d_{\text{sem}}}$ and $\hat{z}_{\text{pix}} =\mathcal{E}_\text{pix}(x) \in \mathbb{R}^{d_{\text{pix}}}$ are the encoded features from semantic and pixel encoder.

\textbf{Quantization.}
We introduce an innovative quantization approach that employs dual codebooks: semantic-level embeddings $\mathbf{Z}_{\text{sem}} = \{z_{\text{sem},i}\}_{i=1}^K \in \mathbb{R}^{K \times d_{\text{sem}}}$  and pixel-level embeddings $\mathbf{Z}_{\text{pix}} = \{z_{\text{pix},i}\}_{i=1}^K \in \mathbb{R}^{K \times d_{\text{pix}}}$, where $K$ is the number of codebook entries. These two codebooks share a unified mapping, enabling simultaneous consideration of high-level semantic information and low-level pixel details during the quantization process. 
Given the encoded feature representations $\hat{z}_{\text {sem}}$ and $\hat{z}_{\text {pix}}$, we compute the distances to their respective codebook embeddings after $l_2$-norm \citep{vitvqgan}:

\begin{equation}
d_{\text{sem},i} = \|\hat{z}_{\text{sem}} - z_{\text{sem},i}\|_2^2, \text{for } i = 1, \ldots, K
\end{equation}

\begin{equation}
d_{\text{pix},i} = \|\hat{z}_{\text{pix}} - z_{\text{pix},i}\|_2^2, \text{for } i = 1, \ldots, K
\end{equation}

\begin{equation}
i^* = \argmin_i (d_{\text{sem},i} + w_\text{dis} \cdot d_{\text{pix},i})
\label{argmin_dis}
\end{equation}

The optimal quantization index $i^*$ is determined by minimizing the weighted sum of these two distances, where $w_\text{dis}$ is the distance balance weight, as shown in \cref{argmin_dis}. 
This joint optimization approach differs significantly from previous VQ methods that typically focus on learning the distribution of a single feature type. 
We further adopt the multi-scale VQ (MSVQ) structure \cite{var} to to enhance the richness of the codebook representation.
Our shared mapping strategy enables the codebook to learn the joint distribution of high-level semantics and low-level features, resulting in several key advantages:

\ding{182} \textit{Scalability}:
Our approach demonstrates consistent performance improvements in both generative and understanding tasks as the codebook size increases, since large codebook size offers more high- and low-level feature combination possibilities. With an expanded codebook size of 131,072, it can still maintain a remarkably high utilization rate of over 95\% while achieving best image reconstruction quality and multimodal understanding performance.

\ding{183} \textit{Multi-task Capabilities}: By learning the joint distribution of semantic and pixel-level features, our method bridges the gap between generation and understanding tasks. This unified representation enables a single tokenizer to excel in both domains. This design also allows seamless integration of more codebooks to embed other type of feature representations, enabling extensibility to more downstream tasks without architectural modifications.

\textbf{Decoder and Training Objective.} 
Our architecture incorporates two distinct decoders, including semantic decoder $\mathcal{D}_\text{sem}$ and pixel decoder $\mathcal{D}_\text{pix}$ for reconstructing semantic features and original image.
We employ a teacher model \cite{peng2022beit} (identical to the semantic encoder's initialization) for target feature extraction. The semantic loss $\mathcal{L}_\text{sem}$ is computed as the $l_2$ distance between decoded and teacher-extracted features. The reconstruction loss is formulated as:
\begin{equation}
\mathcal{L}_{\text{pix}} = {\ell}_2(x, \hat{x}) + \mathcal{L}_{\text{P}}(x, \hat{x}) + \lambda_{\text{G}} \mathcal{L}_{\text{G}}(\hat{x})
\end{equation}
where $\hat{x} = \mathcal{D}_\text{pix}(z)$, ${\ell}_2$ represents pixel-wise reconstruction loss, $\mathcal{L}_{\text{P}}(\cdot)$ denotes perceptual loss using LPIPS, and $\mathcal{L}_{\text{G}}(\cdot)$ represents adversarial loss with $\lambda_{\text{G}}$ as its weight coefficient. Following vector quantization conventions, we employ a straight-through gradient estimator: $z = \text{sg}[z - \hat{z}] + \hat{z}$
where $\text{sg}[\cdot]$ denotes the stop-gradient operation.
The codebook learning objective is:
$\mathcal{L}_{\text{VQ}} = ||\text{sg}[\hat{z}] - z||_2^2 + \beta||\hat{z} - \text{sg}[z]||_2^2 $
where the second term represents commitment loss with balancing factor $\beta$. The total training objective is the sum of all losses:
$
\mathcal{L}_{\text{total}} = \mathcal{L}_{\text{sem}} + \mathcal{L}_{\text{VQ}} + \mathcal{L}_{\text{pix}}
$.

\begin{figure}[htp]
\centering
\includegraphics[width=1.0\linewidth]{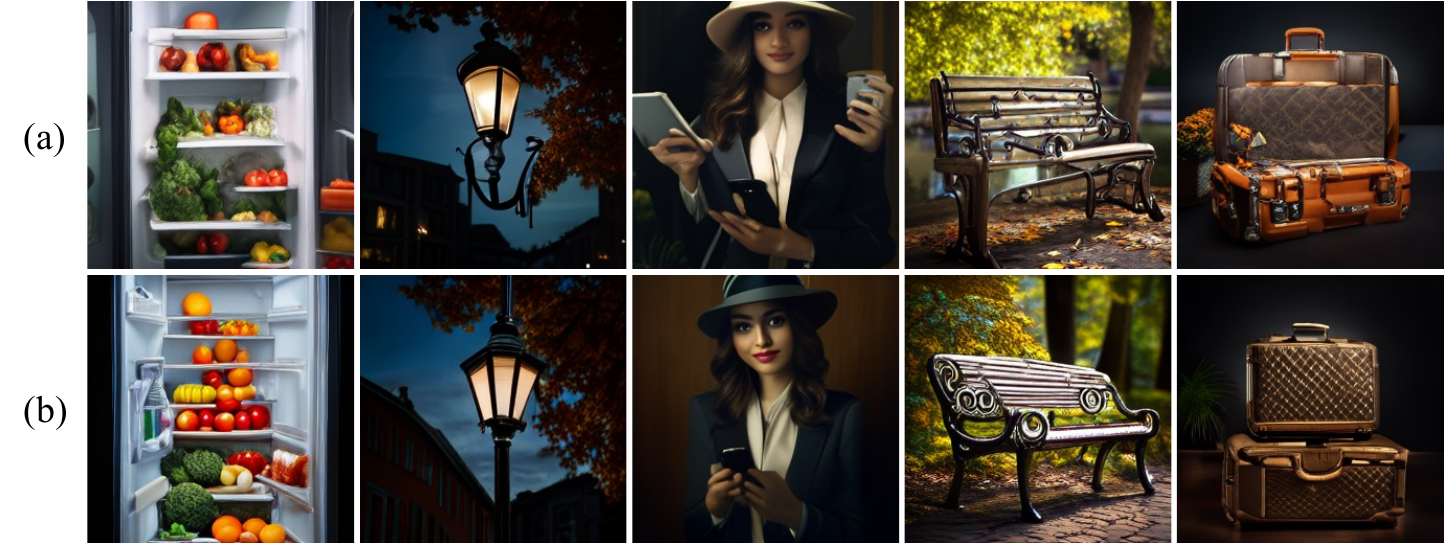}
\caption{Qualitative comparison of different sampling strategies in our framework. (a) Single-pass top-$k$ ($k$=1200) and top-$p$ ($p$=0.8) sampling exhibits inconsistent patterns and artifacts. (b) Our proposed multi-step sampling strategy produces more coherent and visually appealing results. Best zoomed in for details.}
\label{fig:sampling_strategy}
\end{figure}

\subsection{Visual Generation with TokenFlow}
\label{3.3}

TokenFlow helps us achieve SOTA performance in autoregressive text-to-image generation using the next-scale prediction paradigm. Below, we detail our training and inference strategy for high-quality image synthesis.

\textbf{Training Strategy.}
Our visual generation architecture builds upon a pre-trained LLM model~\cite{llama2}. For text encoding, we leverage the model's native BPE tokenizer to transform input text into discrete token sequences and extract feature representations. The original vocabulary is extended with specialized visual tokens. We extract the image tokens using TokenFlow, pass it through a MLP, and concatenate it with text tokens for training.
Given the model's autoregressive nature, we employ cross-entropy loss computed exclusively on image tokens. To enable classifier-free guidance \cite{cfg} during inference, we randomly replace conditioned text with an empty string with probability $p_{\text{drop}}=0.1$ during training. Following \cite{team2024chameleon, dehghani2023scaling, wortsman2023small}, we incorporate QK-normalization and norm re-ordering to enhance training stability and prevent loss spikes.

\textbf{Inference Strategy.}
We observed that conventional top-$k$-top-$p$ sampling strategies, when employed in the next-scale paradigm, often lead to image collapse and repetitive local patterns. This can be attributed to the cross-entropy training objective, which establishes attention-based relationships primarily with the top-1 prediction. Independent top-$k$ sampling for each token during inference can result in tokens lacking direct correlations, leading to inconsistent or repetitive patterns that can only be partially remedied through subsequent scales' attention. This issue becomes more severe particularly with limited inference steps.

To address this fundamental limitation, we propose a novel multi-step sampling approach:
(i) Initial sampling: Perform top-$k$ top-$p$ sampling with parameters $k_1$ and $p_1$. (ii) Refinement: Use the sampled output as input for a second round of sampling in the same scale with reduced parameters $k_2<k_1$ and $p_2<p_1$. 
This progressive narrowing of the sampling space maintains creative diversity while enforcing consistency through refinement steps. Empirical results demonstrate significantly more coherent and visually appealing generations compared to single-pass sampling methods (see \cref{fig:sampling_strategy} and detailed ablation in \cref{additional_ablation_study}).

\subsection{Multimodal Understanding with TokenFlow}
\label{3.4}

TokenFlow functions as a multi-scale VQ tokenizer, where the quantized multi-scale features can be directly fed into a pre-trained LLM for multimodal understanding training, following the LLaVA-1.5 \cite{llava} paradigm. The joint feature representations from dual flow serve as input to the model.
We validate multiple feature input strategies: \textit{(i)} Feature from all scales \textit{(ii)} Final-scale feature only \textit{(iii)} Residual features from all scales. We discover that features from the final scale achieves best overall performance, as detailed in \cref{additional_ablation_study}. This suggests that the final scale captures the most relevant semantic information for multimodal understanding, while additional scale features or residual features may introduce noise that compromises performance. 
Our model demonstrates substantial improvements over existing discrete multimodal methods. Notably, the performance gains can be achieved with minimal computational overhead, requiring less than 24 hour training on 8×A100 GPUs using LLaVA 1.5 training data. 
\section{Experiments}
\subsection{Experimental Setup}
\label{subsec:4.1}

\textbf{Datasets.}
TokenFlow is trained on LAION \cite{laion5b} and COYO-700M \cite{kakaobrain2022coyo700m} and evaluate it on ImageNet \cite{deng2009imagenet}. To enhance face generation quality, we follow \cite{team2024chameleon} and upsample the percentage of images with faces during tokenizer training by 2 times.
For ablation studies, we train the tokenizer for 50 epochs on ImageNet-1K with CLIP ViT-B/14-224 \cite{clip}.
For visual generation with TokenFlow, we trained it on a curated dataset of 60M high-quality images, with captions generated using Qwen-VL \cite{qwenvl}.

\textbf{Implement Details.}
We employ three variants of TokenFlow (B/L/XL), using CLIP ViT-B/14-224 \cite{clip}, ViTamin-XL-256 \cite{chen2024vitamin}, and SigLIP-SO400M-patch14-384 \cite{siglip} as respective teacher models and semantic encoder initializations. Detailed configurations are provided in \cref{appendix:training_details}. For multimodal understanding, we employ Vicuna-v1.5-13B \cite{chiang2023vicuna} and Qwen-2.5-14B \cite{qwen2.5} as the language backbone. 
For 256×256 visual generation training, we truncate captions to first sentence with 0.2 probability to enhance short prompt generation capabilities. The model is initialized with Llama-2-7b \cite{llama2}, and being trained for 2 epochs. At inference, we apply classifier-free guidance \cite{cfg} with a scale factor of 7.5.

\textbf{Evaluation Metrics.} We assess reconstruction quality using rFID, PSNR, and SSIM on the ImageNet-1K validation set \cite{deng2009imagenet}. For multimodal understanding, we evaluate on a comprehensive suite of vision-language benchmarks: SEEDBench \cite{seedbench}, MMVet \cite{mmvet}, POPE \cite{pope}, VQAv2 \cite{vqav2}, GQA \cite{gqa}, TextVQA \cite{textvqa}, AI2D \cite{ai2d}, RealWorldQA \cite{realworldqa}, MMMU \cite{yue2024mmmu}, MMBench \cite{liu2025mmbench}, and MME \cite{fu2023mme}.
Visual generation capabilities are evaluated using GenEval \cite{geneval} and DPG-Bench \cite{dpgbench}. We opt not to include FID scores as argued that it does not correlate well with human assessment of the overall performance of generative models \cite{podell2023sdxl,emu1,chen2023pixart}.

\subsection{Unified Image Tokenizer}
\label{subsec:4.2}

\begin{table}[h]
    \centering
    \caption{Comparison of reconstruction quality on the ImageNet 50k validation set. “\#Lvls.” represents the number of residual levels used. For 384×384 resolution, the downsample ratio of 14.2 is derived from 384/27.}
    \label{tab:tokenizer}
    \definecolor{lightblue}{RGB}{240,248,255}
    \newcommand{\colorrow}[1]{\rowcolor{lightblue} #1}
    \resizebox{1.0\linewidth}{!}{
        \begin{tabular}{l|ccc|ccc}
            \toprule
            Model & Res. & ratio & \#Lvls. & rFID $\downarrow$ & PSNR $\uparrow$ & SSIM $\uparrow$ \\
            \midrule
            VQ-GAN \cite{vqgan} & 256 & 16 & 1 & 4.98 & 20.00 & 0.629 \\
            LlamaGen \cite{llamagen} & 256 & 16 & 1 & 2.19 & 20.79 & 0.675 \\
            RQ-VAE \cite{rqvae} & 256 & 32 & 4 & 3.20 & -- & -- \\
            RQ-VAE \cite{rqvae} & 256 & 16 & 4 & 1.30 & -- & -- \\
            VAR \cite{var} & 256 & 16 & 10 & 1.00 & 22.63 & 0.755 \\
            VILA-U \cite{vilau} & 256 & 16 & 4 & 1.80 & -- & -- \\
            \colorrow{Ours & 256 & 16 & 9 & 1.37 & 21.41 & 0.687 \\}
            \midrule
            LlamaGen \cite{vilau} & 384 & 14.2 & 1 & 0.94 & 21.94 & 0.726 \\
            VILA-U \cite{vilau} & 384 & 14.2 & 16 & 1.25 & -- & -- \\
            VAR \cite{var} & 384 & 16 & 13 & 2.09 & 22.73 & \textbf{0.774} \\
            \colorrow{Ours & 384 & 14.2 & 15 & \textbf{0.63} & \textbf{22.77} & 0.731 \\}
            \bottomrule
        \end{tabular}
    }
\end{table}

In \cref{tab:tokenizer}, we present reconstruction metrics of TokenFlow on 256×256 and 384×384 resolutions. The metric of VAR \cite{var} is tested with the released checkpoint. At 256×256 resolution with a 16× compression ratio, TokenFlow achieves competitive performance with an rFID of 1.37, comparable to RQ-VAE while significantly outperforming previous methods such as VQ-GAN and LlamaGen. TokenFlow demonstrates superior reconstruction quality across all metrics in 384×384 resolution—a standard size in multimodal understanding tasks. These results validate the effectiveness of dual codebook design in preserving fine-grained visual details. Moreover, the incorporation of shared mapping enables TokenFlow to maintain high-level semantic features, as verified in \cref{subsec:4.3}.

\subsection{Multimodal Understanding}
\label{subsec:4.3}

\begin{table*}[h]
    \centering
    \caption{Evaluation on multimodal understanding benchmarks. We collect evaluations including:  
        SEEDB: SEED Bench-Img~\cite{seedbench};
        MMV: MM-Vet~\cite{mmvet};
        POPE~\cite{pope}; 
        VQAv2~\cite{vqav2}; 
        GQA~\cite{gqa};  
        TQA: TextVQA~\cite{textvqa};   
        AI2D~\cite{ai2d};
        RWQA: RealWorldQA~\cite{realworldqa};
        MMMU~\cite{yue2024mmmu};
        MMB: MMBench~\cite{liu2025mmbench}; MME \cite{fu2023mme} and MME-P: MME-Perception. We include approaches with continuous visual inputs (top) versus discrete visual inputs (bottom). The best results among approaches with discrete visual input are highlighted in bold. * results are not reported in original paper and tested with lmms-eval \cite{lmms_eval2024} using the released checkpoint. When calculating average, we use MME-P and divide it by 20 to have the same scale with other benchmarks.
    }

    \label{tab:und_results}

    \definecolor{lightblue}{RGB}{240,248,255}
    \newcommand{\colorrow}[1]{\rowcolor{lightblue} #1}
    
    \resizebox{\linewidth}{!}{
        \begin{tabular}{l | cc | ccc cccc cc ccc cc | c}
            \toprule
            Method & \# Params & Res.
            & SEEDB & MMV & POPE
            & VQAv2 & GQA & TQA 
            & AI2D & RWQA & MMMU & MMB
            & MME & MME-P & Avg.
            \\
            \midrule
            \multicolumn{16}{l}{\textit{Continuous Visual Input}}\\
            \midrule
            InstructBLIP \cite{instructblip}
            & Vicuna-13B & 224
            & 58.8 & 25.6 & 78.9 
            & --   & 49.5 & 50.7 
            & -- & -- & -- & 36.0 
            & -- & 1212.8 & --
            \\
            MiniGPT-4 \cite{minigpt4}
            & Vicuna-13B & 224
            & -- & -- & -- 
            & --   & -- & -- 
            & -- & -- & -- & -- 
            & 1158.7 & 866.6 & --
            \\
            BLIP-2 \cite{blip2}
            & Vicuna-13B & 224
            & 46.4 & 22.4 & -- 
            & --   & -- & 42.5 
            & -- & -- & 26.6 & -- 
            & -- & 1293.8 & --
            \\
            ShareGPT4V \cite{chen2023sharegpt4v}
            & Vicuna-7B & 336
            & 69.7 & 37.6 & --
            & 80.6 & 63.3 & 60.4 
            & 58.0 & 54.9 & 37.2 & 68.8 
            & 1943.8 & 1567.4 & --
            \\
            NExT-GPT \cite{nextgpt} 
            & Vicuna-7B & 224
            & 57.5 & -- & --
            & 66.0 & --   & -- 
            & --   & --   & -- & 58.0
            & -- & -- & --
            \\
            Qwen-VL-Chat \cite{qwenvl} 
            & Qwen-7B & 448
            & 57.7 & -- & --
            & 78.2 & 57.5   & -- 
            & --   & --   & -- & --
            & 1848.3 & 1487.5 & --
            \\
            Janus~\cite{wu2024janus} 
            & DeepSeek-LLM-1.3B & 384
            & 63.7 & 34.3 & 87.0
            & 77.3 & 59.1 & -- 
            & --   & --   & 30.5 & 69.4
            & -- & 1338.0 & --
            \\
            LLaVA-1.5 \cite{llava}
            & Vicuna-13B & 336
            & 68.1 & 36.1 & 85.9	
            & 80.0 & 63.3 & 61.3 
            & 61.1 & 55.3 & 36.4 & 67.7 
            & 1826.7 & 1531.3 & 62.9
            \\
            \midrule
            \multicolumn{16}{l}{\textit{Discrete Visual Input}}
            \\
            \midrule
            Gemini-Nano-1~\cite{team2023gemini}
            & 1.8B from scratch & --
            & --   & --   & --
            & 62.7 & --   & -- 
            & --   & --   & 26.3 & --  
            & -- & -- & --
            \\
            Chameleon~\cite{team2024chameleon}
            & 34B from scratch & 256
            & --   & --   & --
            & 69.6 & --   & -- 
            & --   & --   & -- & --  
            & -- & -- & --
            \\
            LWM~\cite{lwm}
            & LLaMA-2-7B & 256
            & --   & 9.6 & 75.2
            & 55.8 & 44.8 & 18.8 
            & --   & --   & -- & --  
            & -- & -- & --
            \\
            SEED-LLaMA~\cite{li2024seed}
            & LLaMA-2-13B & 224
            & 53.7 & -- & --
            & 63.4 & -- & -- 
            & --   & --   & -- & --  
            & -- & -- & --
            \\
            Show-o~\cite{showo} 
            & Phi-1.5-1.3B & 256
            & -- & --   & 80.0
            & 69.4 & 58.0 & -- 
            & --   & --   & 26.7 & --  
            & -- & 1097.2 & --
            \\
            VILA-U~\cite{vilau} 
            & LLaMA-2-7B    & 256
            & 56.3  & 27.7  & 83.9
            & 75.3  & 58.3  & 48.3 
            & --    & --    & --    & --  
            & --    & 1336.2 & --
            \\
            VILA-U~\cite{vilau} 
            & LLaMA-2-7B    & 384
            & 59.0  & 33.5  & 85.8
            & \textbf{79.4}  & 60.8  & 60.8 
            & --    & --    & --    & --  
            & --    & 1401.8 & --
            \\
            EMU3 \cite{wang2024emu3}
            & 8B from scratch   & 512 
            & 68.2  & 37.2  & 85.2
            & 75.1  & 60.3  & \textbf{64.7}
            & 70.0 & \textbf{57.4} & 31.6 & 58.5
            & 1509.9*    & 1243.8* & 60.9
            \\
            \colorrow{TokenFlow-B
            & Vicuna-13B   & 224
            & 60.4  & 22.4  & 84.0
            & 70.2  & 59.3  & 49.8
            & 54.2  & 49.4  & 34.2  & 55.3
            & 1660.4    & 1353.6 & 55.2
            \\}
            \colorrow{TokenFlow-L
            & Vicuna-13B   & 256
            & 62.6  & 27.7  & 85.0
            & 73.9  & 60.3  & 54.1
            & 56.6  & 49.2  & 34.4  & 60.3
            & 1622.9    & 1365.4 & 57.5
            \\}
            \colorrow{TokenFlow-XL
            & Vicuna-13B   & 384 
            & 68.7     & 40.7     & 86.8
            & 77.9    & \textbf{62.7}     & 61.5
            & 66.7  & 53.7  & 38.7     & 68.9
            & 1840.9   & 1545.9 & 64.0
            \\}
            \colorrow{TokenFlow-XL
            & Qwen-2.5-14B   & 384 
            &  \textbf{72.6}     &  \textbf{48.2}     & \textbf{87.8}
            &  77.6     &  62.5     & 62.3
            &  \textbf{75.8}     &  56.6     &  \textbf{43.2}    & \textbf{76.8}
            &  \textbf{1922.2}     & \textbf{1551.1}     & \textbf{67.4}
            \\}
            \bottomrule
            \end{tabular}
    }
\end{table*}

TokenFlow, as a discrete visual encoder,  demonstrates state-of-the-art performance across a comprehensive suite of multimodal understanding benchmarks. Following LLaVA-1.5's training pipeline, we train TokenFlow-B and TokenFlow-L using LLaVA-Pretrain558K for adapter pretraining and LLaVA-v1.5-mix-665K for instruction tuning. For TokenFlow-XL, inspired by recent findings in \cite{tong2024cambrian}, we leverage Cambrian-Alignment and Cambrian-10M for pretraining and instruction tuning respectively, as the teacher model SigLIP-SO400M benefits significantly from increased training data.
As evidenced in \cref{tab:und_results}, TokenFlow-XL achieves competitive or superior results compared to leading approaches with continuous inputs from CLIP-style encoders. Using the same language backbone (Vicuna 13B), TokenFlow-XL outperforms LLaVA-1.5 13B by 1.7\% on average, for the first time demonstrates that model with discrete visual input can surpass this strong baseline.
By simply changing the LLM backbone to Qwen-2.5-14B \cite{qwen2.5}, we further surpass LLaVA-1.5 by 7.2\%.

When compared to methods using discrete inputs, our approach demonstrates superior performance while maintaining training efficiency. Unlike models trained from scratch such as Chameleon and EMU3, our method requires less than 24 hour of training on 8×A100 GPUs using LLaVA 1.5 data. 
TokenFlow-XL 14B significantly outperforms EMU3 with an overall improvement of 10.7\%.
Given these promising empirical results, we position TokenFlow as a potential next-generation vision tokenizer for unified understanding and generation tasks. Our findings suggest that discrete visual representations can not only match but exceed the performance of continuous counterparts while maintaining practical training requirements.

\subsection{Visual Generation}
\label{subsec:4.4}

We evaluate our model's generation capabilities against state-of-the-art methods including diffusion-based, autoregressive-based, and hybrid approaches on standard benchmarks GenEval \cite{geneval} and DPG-Bench \cite{dpgbench}. As shown in \cref{tab:gen_results}, our approach achieves competitive performance while requiring significantly fewer generation steps.

For 256×256 image generation, we employ a multi-step sampling strategy instead of the original 9-step sampling (one per tokenizer scale). Specifically, we apply three steps per scale with top-$k$=\texttt{[1200,100,1]} and top-$p$=\texttt{[0.8,0.8,1.0]} across all scales except the first, totaling 25 steps.
Under this inference scheme, our model achieves a GenEval score of 0.55, surpassing prominent diffusion models like Stable Diffusion v2.1 and PixArt-alpha. More significantly, it surpasses autoregressive methods such as Chameleon, LlamaGen, and EMU3, which require thousands of inference steps. With prompt rewriting, our model achieves 0.63, approaching DALL-E 3's performance.
On DPG-Bench, it achieves an average score of 72.9, outperforming LlamaGen, Show-o, SD v1.5, and PixArt-alpha.
Moreover, our model only requires 2.7 seconds to infer one image with 1×A100 GPU, which is significantly faster than other autoregressive-based methods. 

\begin{table}[htp]
    \centering
    \caption{Comparison of generation quality on GenEval \cite{geneval} and DPG-Bench \cite{dpgbench}. "\#Step": the number of model runs needed to generate an image. $\dagger$ result is with rewriting.}

    \label{tab:gen_results}
    \definecolor{lightblue}{RGB}{240,248,255}
    \definecolor{lightgray}{RGB}{229,229,229}
    \definecolor{customgray}{RGB}{200,200,200}

    \newcommand{\bluerow}[1]{\rowcolor{lightblue} #1}
    \newcommand{\grayrow}[1]{\rowcolor{lightgray} #1}
    \resizebox{1.0\linewidth}{!}{
        \begin{tabular}{l | ccc | ccc}
            \toprule
            \multirow{2}{*}{Model} & \multirow{2}{*}{Text Pretrain} & \multirow{2}{*}{Res.} & \multirow{2}{*}{\#Steps} & GenEval & DPG-Bench \\
            & & & & Overall $\uparrow$ & Average $\uparrow$ \\
            \midrule
            \multicolumn{6}{l}{\textit{Diffusion-based}}\\
            \midrule
            SD v1.5 \cite{sd}     & CLIP ViT-L/14 & 512 & 50 & 0.43 & 63.18 \\
            DALL-E 2 \cite{dalle2}   & CLIP ViT-H/16 & 1024 & -- & 0.52 & -- \\
            SD v2.1 \cite{sd}     & CLIP ViT-H/14 & 768 & 50 & 0.50 & -- \\
            SDXL  \cite{podell2023sdxl}      & CLIP ViT-bigG & 1024 & 40 & 0.55 & 74.65 \\
            PixArt-alpha \cite{chen2023pixart}   & Flan-T5-XXL & 512 & 20 & 0.48 & 71.11 \\
            DALL-E 3 \cite{dalle3}       & Flan-T5-XXL & 1024 & -- & 0.67$^\dagger$ & 83.50 \\
            \midrule
            \multicolumn{6}{l}{\textit{Autoregressive meets diffusion}}\\
            \midrule
            Show-o \cite{showo} & Phi-1.5 & 256 & 16 & 0.53 & 67.27 \\
            Transfusion \cite{zhou2024transfusion} & -- & 256 & 250 & 0.63 & -- \\
            \midrule
            \multicolumn{6}{l}{\textit{Autoregressive-based}}\\
            \midrule
            Chameleon \cite{team2024chameleon} & -- & 512 & 1024 & 0.39 & -- \\
            LlamaGen \cite{llamagen} & Flan-T5-XL & 512 & 1024 & 0.32 & 64.84 \\
            EMU3 \cite{wang2024emu3} & -- & 512 & 4096 & 0.54 \textcolor{customgray}{/ 0.66$^\dagger$}  & 80.60 \\
            \grayrow{VAR \cite{var} & -- & 256 & 28 & 0.53 & 71.08 \\}
            \bluerow{Ours & -- & 256 & 25 &  0.55 \textcolor{customgray}{/ 0.63$^\dagger$} & 73.38 \\}
            \bottomrule
        \end{tabular}
    }
\end{table}

\begin{figure}[htp]
\centering
\includegraphics[width=1.0\linewidth]{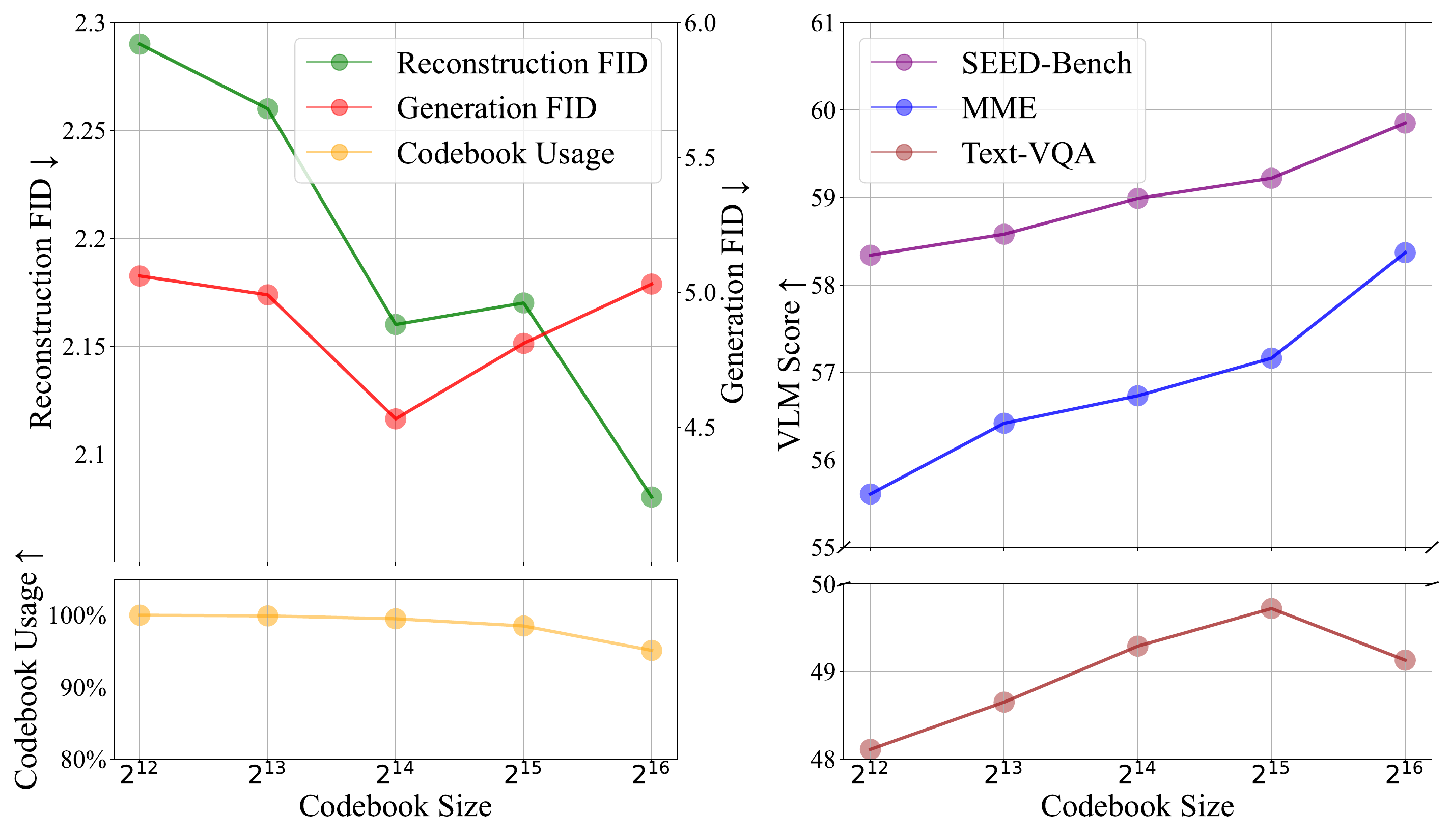}
\vspace{-2em}
\caption{Impact of codebook size on reconstruction quality, class-conditional generation, and multimodal understanding benchmarks. MME is divide by 28 to have the same scale.}
\label{fig:ablation_key_design}
\vspace{-1em}
\end{figure}

We further conduct additional text-to-image comparison between TokenFlow and the released VAR tokenizer \cite{var}. Under identical training configurations and dataset settings, our model consistently demonstrates better performance across all benchmark metrics, this further showcasing the effectiveness of our unified tokenization approach.

\subsection{Ablation Studies}
\label{ablation}

\textbf{Effect of Codebook Size.}
In \cref{fig:ablation_key_design}, we experimented the impact of codebook size in our unified tokenizer, varying from 8,192 to 131,072. Our evaluation spans reconstruction quality, class-conditional generation, and multimodal understanding capabilities. For class-conditional generation, we employ the VAR transformer \cite{var} with d=16, resulting in approximately 310M parameters.
\begin{table}[htp]
    \centering
    \caption{Impact of key design choices on reconstruction quality and multimodal understanding benchmarks. Best results for each metric are highlighted in bold.}
    \label{tab:ablation_key_design}
    \resizebox{1.0\linewidth}{!}{
        \begin{tabular}{ccc|cccc}
        \toprule
        Shared Mapping & MSVQ & CLIP Init. & rFID $\downarrow$ & MME-P $\uparrow$ & SEEDB $\uparrow$ & TQA $\uparrow$ \\
        \midrule
        
        & & & 8.07 & 1252.38 & 57.84 & 49.16 \\
        \usym{2714} & & & 3.96 & 1212.51 & 55.97 & 47.42 \\
        \usym{2714} & \usym{2714} & & 2.18 & 1209.90 & 56.08 & 47.40 \\
        \usym{2714} & \usym{2714} & \usym{2714} & \textbf{2.16} & \textbf{1312.09} & \textbf{58.99} & \textbf{49.29} \\
        \bottomrule
        \end{tabular}
    }
\end{table}
Notably, our approach maintains a consistently high codebook utilization rate exceeding 95\% even with codebook size of 131,072, attributed to our shared mapping design. The shared mapping allows for effective combinations of high-level semantic features and low-level details, addressing a common limitation of conventional VQ tokenizers \cite{vqgan} that typically suffer from deteriorating utilization rates at larger scales.

Our results reveal that increasing codebook size enhances performance across multimodal understanding benchmarks and reconstruction quality.
However, when codebook size exceeds 32,768, we observe a slight degradation in class-conditional generation performance. This phenomenon can be attributed to the increased complexity of learning for autoregressive generation with larger codebooks. Based on this finding, we adopt a codebook size of 32,768 for our text-to-image generation experiments.

\textbf{Effect of Key Design Choice.}
We validate the effectiveness of our key design choices in TokenFlow: shared mapping, multi-scale vector quantization (MSVQ), and CLIP initialization for the semantic encoder. As shown in \cref{tab:ablation_key_design}, we start with a baseline that uses one single codebook distilled from CLIP ViT-B/14, coupled with a pixel decoder for direct image reconstruction from semantic features. This baseline yields a high reconstruction FID of 8.07, primarily due to the challenge of reconstructing fine-grained pixel details solely from semantic features, as visualized in \cref{fig:reconstruction-from-beitv2}.
The introduction of shared mapping (Row 2) enables the two codebooks to capture high-level and low-level features simultaneously. By weighted distance computation, we quantize the input with optimal combinations of high-level and low-level features. This design significantly improves reconstruction quality (-4.11 rFID) while maintaining comparable understanding capabilities.

We further find that incorporating MSVQ \cite{var} (Row 3) introduces multi-granular information into the codebook embeddings, which results in enhanced reconstruction performance, with rFID of 2.18. Moreover, this hierarchical design enables a next-scale prediction paradigm in downstream text-to-image generation tasks, offering significant inference speed advantages over traditional next-token prediction approaches \cite{var,tang2024hart}.
Initializing the semantic encoder with pretrained CLIP weights (Row 4) while making it unfrozen during tokenizer training provides strong semantic priors for codebook embeddings. This results in substantial improvements across all understanding metrics (+8.4\% in MME-Perception, +5.2\% in SEED-Bench, and +4.0\% in TextVQA).
Given these empirical results, we adopt this configuration as our final model architecture and extend our experiments with stronger teacher models, additional training data, and longer training iterations.

\section{Conclusion}

In this work, we introduce TokenFlow, a novel unified image tokenizer that effectively bridges the gap between multimodal understanding and generation through its innovative dual-codebook architecture. 
By decoupling semantic and pixel-level feature learning while maintaining their alignment via shared mapping, TokenFlow successfully addresses the fundamental issue between different granularities of visual information required for understanding and generation tasks.
Our comprehensive experiments demonstrate its effectiveness across multiple dimensions: superior reconstruction quality at different resolutions, state-of-the-art performance in multimodal understanding with minimal training costs, and competitive visual generation capabilities with substantially fewer inference steps. 
These results validate that decoupled yet aligned feature learning through our shared mapping can effectively unify understanding and generation while maintaining superior performance in both domains, suggesting TokenFlow as a promising next-era foundation tokenizer for vision-language systems.

{
    \small
    \bibliographystyle{ieeenat_fullname}
    \bibliography{main}
}

\clearpage
\setcounter{page}{1}
\maketitlesupplementary
\appendix

\section{Implementation Details}

\subsection{Motivation} 
\label{appendix:1}

\textbf{Experimental Setup for Multimodal Understanding.} 
To evaluate the multimodal understanding capabilities of current VQ tokenizers, we conduct experiments as detailed in \cref{wrap-tab:llava-vq-model}. For LFQ \citep{lfq}, we utilize the open-source implementation \cite{luo2024open}, which demonstrates  comparable performance to the original paper. The codebook size of LFQ is 262,144. For VQGAN-LC \citep{vqgan_lc}, we employ features before its projection layer, which is clustered from the pretrained CLIP image encoder, with a codebook size of 100,000.

\begin{figure}[htp]
    \begin{center}
    \includegraphics[width=1.0\linewidth]{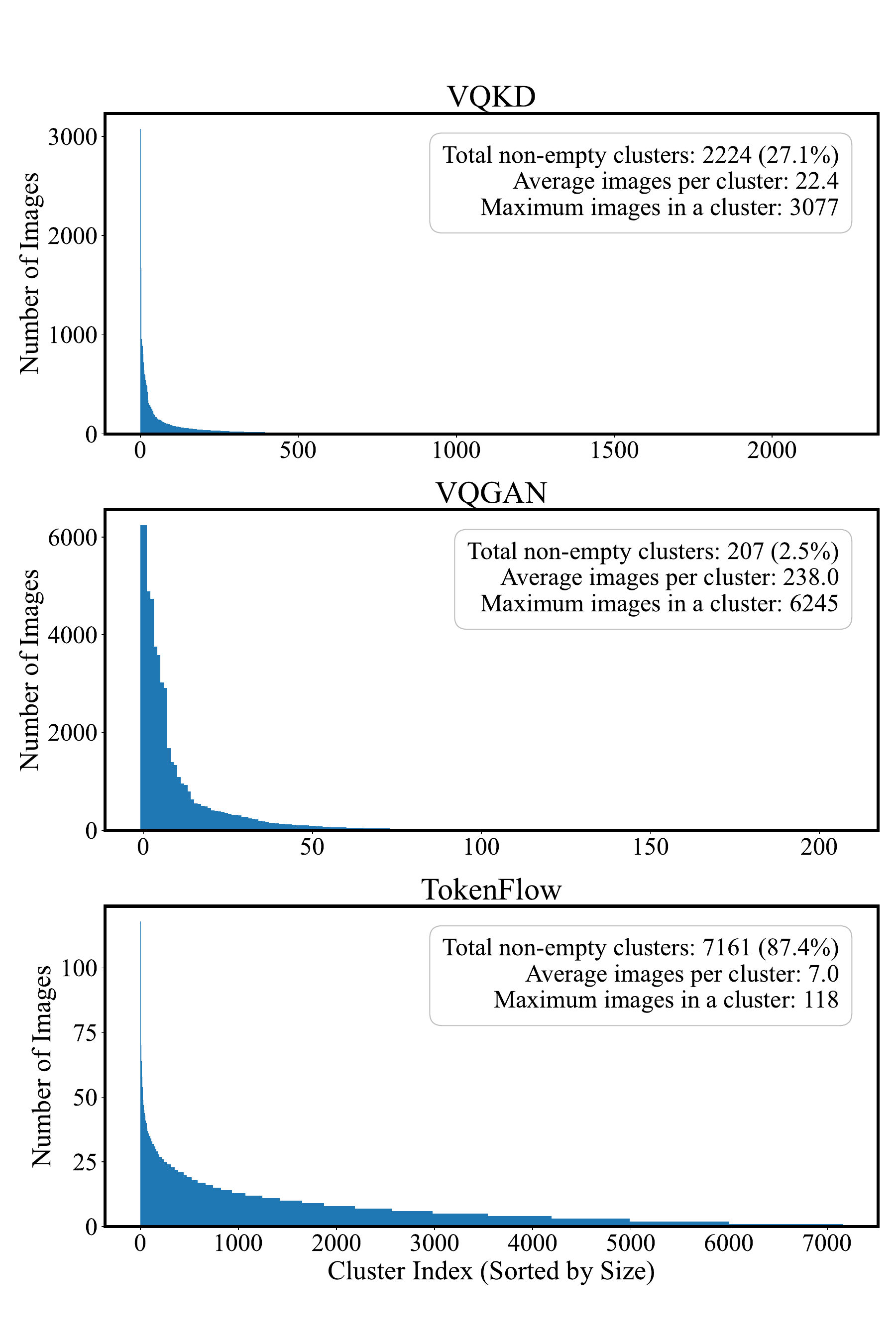}
    \vspace{-3em}
    \caption{Comparison of cluster size distributions between VQKD \cite{peng2022beit}, VQGAN \cite{vqgan}, and TokenFlow (ours), with a fixed codebook size of 8,192. Analysis performed on 50,000 images from the ImageNet-1k validation set. TokenFlow exhibits significantly smoother distribution compared to others, attributed to our shared mapping design that learns joint distributions of semantic and pixel-level features. This joint learning approach helps maintain high codebook utilization (95\%+) even with large-scale codebooks containing over 131K entries.}
    \label{fig:cluster_distribution}
    \end{center}
\end{figure}

\textbf{Experimental Setup for Visual Comparison of VQKD, VQGAN and TokenFlow.} 
To generate the visualizations in \cref{fig:diff-vqgan-beitv2}, we perform an experiment using 50,000 images from the ImageNet-1k validation set. We process these images through the encoders of VQKD, VQGAN and TokenFlow, applying average pooling to the extracted features to obtain a $1 \times 1$ representation. Subsequently, we identify the closest index in their respective codebooks using $l_2$ distance. We provide more visualizations in \cref{fig:cluster3}, and visualize the cluster size distribution in \cref{fig:cluster_distribution}.

\begin{figure*}[htp]
    \begin{center}
    \includegraphics[width=1.0\linewidth]{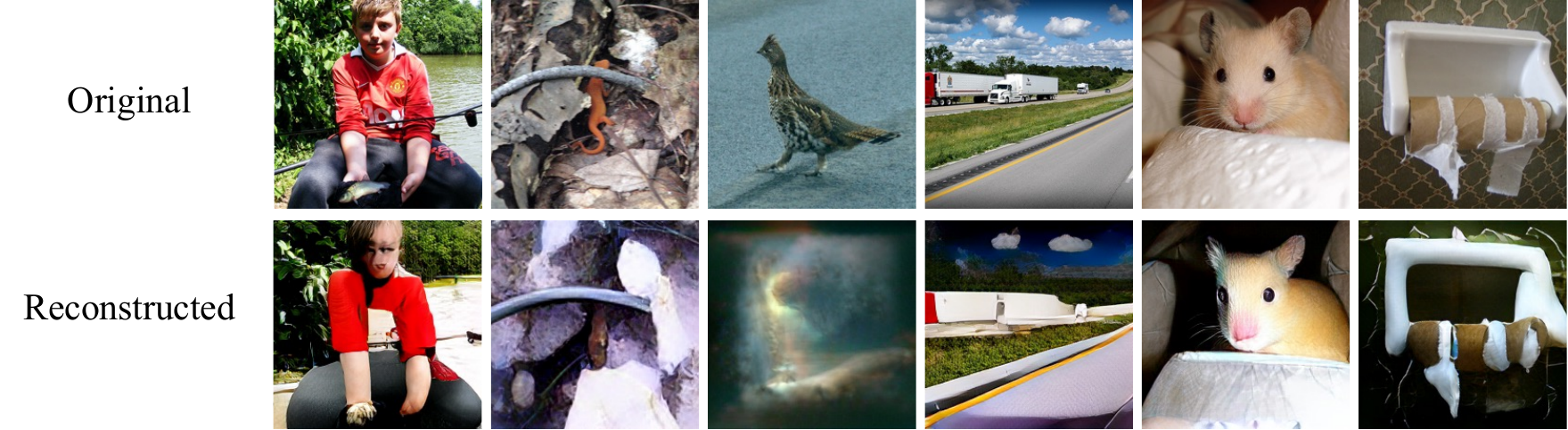}
    \caption{Comparison of original images and their reconstructions from quantized semantic features extracted by VQKD~\cite{peng2022beit}. The reconstructed images preserve the semantic content but exhibit significant loss of high-frequency details.}
    \label{fig:reconstruction-from-beitv2}
    \end{center}
\end{figure*}

\textbf{Experimental Setup for Image Reconstruction from Quantized Semantic Feature.} 
We conducted an experiment to reconstruct original images from quantized features extracted by VQKD \citep{peng2022beit}. In this setup, we maintained the original encoder and quantizer of VQKD, while introducing an additional decoder aimed at reconstructing the input image. The architecture of this decoder is identical to the pixel decoder employed in our TokenFlow. We trained this decoder on the ImageNet-1K dataset for 100 epochs.
\cref{fig:reconstruction-from-beitv2} presents a visual comparison between the original and the reconstructed images. As observed, while the reconstructed images maintain the overall semantic content, they exhibit a noticeable loss of high-frequency details. This phenomenon suggests that the quantized semantic features cannot fully preserve fine-grained visual details, which is crucial for visual generation.

\subsection{Tokenizer Training Details}
\label{appendix:training_details}

We provide detailed training configurations for TokenFlow-B, TokenFlow-L, and TokenFlow-XL variants in \cref{wrap-tab:training_settings}. All models share common hyperparameters including learning rate, batch size, commitment loss factor, adversarial loss factor and distance balance weight. The models primarily differ in their input resolution (224, 256, and 384) and semantic teacher models, utilizing CLIP ViT-B/14 \cite{clip}, ViTamin-XL \cite{chen2024vitamin}, and SigLIP-SO400M \cite{siglip}.

\section{Additional Results}

\subsection{Additional Ablation Study}
\label{additional_ablation_study}

\textbf{Effect of Sampling Strategy to Visual Generation.}
We conduct comprehensive ablation studies to analyze the impact of different sampling strategies on generation quality. As shown in Table \ref{wrap-tab:sampling_geneval_imagereward}, we evaluate various configurations using GenEval \cite{geneval} and ImageReward \cite{xu2024imagereward} metrics. We choose ImageReward for ablation due to its strong correlation with human preferences, particularly in capturing local artifacts and overall visual quality. The ImageReward is average over 10k prompts from the MS-COCO validation set. For multi-step configurations, we denote the top-$p$ and top-$k$ values for each step using bracket notation [$x_1$, ..., $x_n$].

Our multi-step approach with a two-step strategy (top-$k$=[1200, 1], top-$p$=[0.8, 0]) significantly improves generation quality, yielding gains of +0.039 in GenEval and +0.084 in ImageReward compared to single-step sampling. This validates our hypothesis that progressive refinement helps maintain global consistency. When increasing the second-step $k$ value to 10 or 100 while maintaining top-$p$, we observe slightly degraded performance. This degradation suggests that excessive sampling freedom in refinement steps can lead to increased artifacts and local inconsistencies.

Most notably, three-step strategy (top-$k$=[1200, 100, 1], top-$p$=[0.8, 0.8, 0]) achieves the best performance across both metrics. This represents substantial improvements of 10.2\% and 14.3\% over traditional single-step sampling, respectively. The gradual narrowing of sampling space (1200→100→1) strikes a balance between generation diversity and local consistency. As illustrated in Figure \ref{fig:sampling_strategy}, our multi-step approach produces more coherent and visually appealing results.
These quantitative and qualitative results demonstrates that progressive refinement in top-$p$ top-$k$ sampling is crucial for high-quality generation in next-scale prediction frameworks.

\begin{table}[htp]
\small
\begin{center}
\caption{Impact of sampling strategy to visual generation. We compare single-step \textit{v.s.} multi-step sampling strategy using GenEval and ImageReward. For multi-step approaches, values in brackets indicate parameters for successive sampling steps.}
\label{wrap-tab:sampling_geneval_imagereward}
\resizebox{\linewidth}{!}{
\begin{tabular}{lcc|cc}
    \toprule  
    Strategy & Top-\textit{k} & Top-\textit{p} & GenEval $\uparrow$ & ImageReward $\uparrow$ \\
    \midrule
    Single Step & 1200 & 0.8 & 0.502 & 0.722 \\
    \midrule
    \multirow{4}{*}{Multi Step} & [1200, 1] & [0.8, 0] & 0.541 & 0.806 \\
                                  & [1200, 10] & [0.8, 0.8] & 0.531 & 0.799 \\
                                  & [1200, 100] & [0.8, 0.8] & 0.529 & 0.745 \\
                                  & [1200, 100, 1] & [0.8, 0.8, 0] & \textbf{0.553} & \textbf{0.825} \\
    \bottomrule
\end{tabular}
}
\end{center}
\end{table}

\begin{table}[htp]
\small
\begin{center}
\caption{Impact of model size to visual generation.}
\label{wrap-tab:model_size_ablation}
\resizebox{\linewidth}{!}{
\begin{tabular}{cc|cc}
    \toprule  
    Model size & Training epoches & GenEval $\uparrow$ & ImageReward $\uparrow$ \\
    \midrule
    1B & 4 & 0.485 & 0.677 \\
    7B & 2 & \textbf{0.553} & \textbf{0.825} \\
    \bottomrule
\end{tabular}
}
\end{center}
\end{table}

\begin{table}[htp]
\small
\begin{center}
\caption{Impact of different input strategies on multimodal understanding. Best results for each metric are highlighted in bold.}
\label{wrap-tab:input_multimodal_und}
    \definecolor{lightblue}{RGB}{240,248,255}
    \definecolor{lightgray}{RGB}{229,229,229}
    \newcommand{\bluerow}[1]{\rowcolor{lightblue} #1}
    \newcommand{\grayrow}[1]{\rowcolor{lightgray} #1}
    \resizebox{\linewidth}{!}{
    \begin{tabular}{lcccc}
        \toprule  
        Input strategy & MME $\uparrow$ & MME-P $\uparrow$ & SEEDB $\uparrow$ & TQA $\uparrow$ \\
        \midrule
        Full scale & 1610.1 & 1315.1 & 59.6 & 49.5 \\
        Full scale residual & 1527.5 & 1216.5 & 57.0 & 48.1 \\
        Last scale semantic feat. only  & 1580.3 & 1315.6 & \textbf{60.1} & \textbf{49.7} \\
        Last scale  & \textbf{1634.3} & \textbf{1356.5} & 59.9 & 49.1 \\
        \bottomrule
    \end{tabular}
    }
\end{center}
\end{table}

\begin{figure}[htp]
    \begin{center}
    \includegraphics[width=1.0\linewidth]{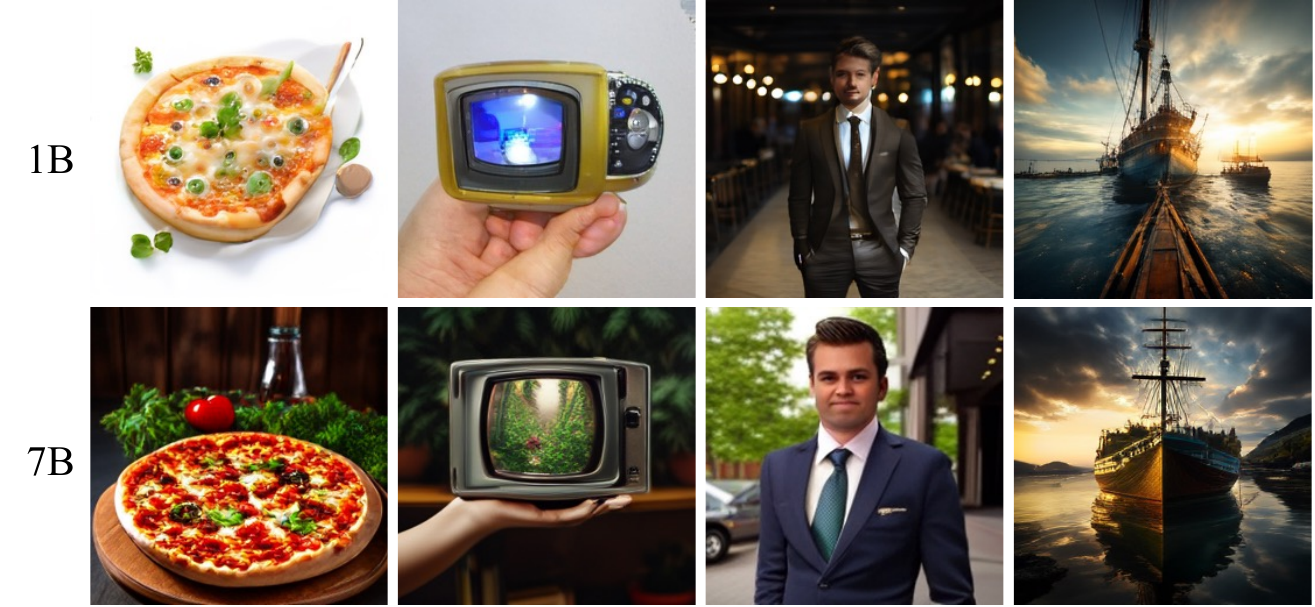}
    \caption{Qualitative comparison of visual generation capabilities between 1B and 7B models. Prompts (from left to right): (1) "A pizza sitting on top of a wooden cutting board", (2) "Television set being held by a hand", (3) "The guy is nicely dressed in a suit and tie", and (4) "A sailing ship rests on waters". The 7B model demonstrates enhanced quality compared to its 1B counterpart.}
    \label{fig:model_size_ablation}
    \end{center}
\end{figure}

\textbf{Effect of Model Size to Visual Generation.}
We conduct ablation studies to investigate the impact of model size on our decoder-only visual generation architecture. Specifically, we initialize our framework with two different backbone models: TinyLlama-1B \cite{zhang2024tinyllama} and Llama-2-7B \cite{llama2}. Experiments demonstrate that model size plays a crucial role in generation performance. As shown in \cref{wrap-tab:model_size_ablation} and \cref{fig:model_size_ablation}, under identical sampling strategies and training dataset configurations, the 1B model significantly underperforms compared to its 7B counterpart, even with doubled training epochs.

\textbf{Effect of Input Strategy to Multimodal Understanding.}
We validate different feature input strategies for multimodal understanding with TokenFlow. As shown in \cref{wrap-tab:input_multimodal_und}, final-scale features consistently outperform both full-scale features and full-scale residual features across all benchmarks. This suggests that the final scale captures the most relevant semantic information for multimodal understanding, while additional scale features or residual features may introduce noise that compromises performance. Our experiments also reveal that utilizing semantic features only does not improve the overall understanding performance.

\begin{figure}[!h]
\centering
\includegraphics[width=1.0\linewidth]{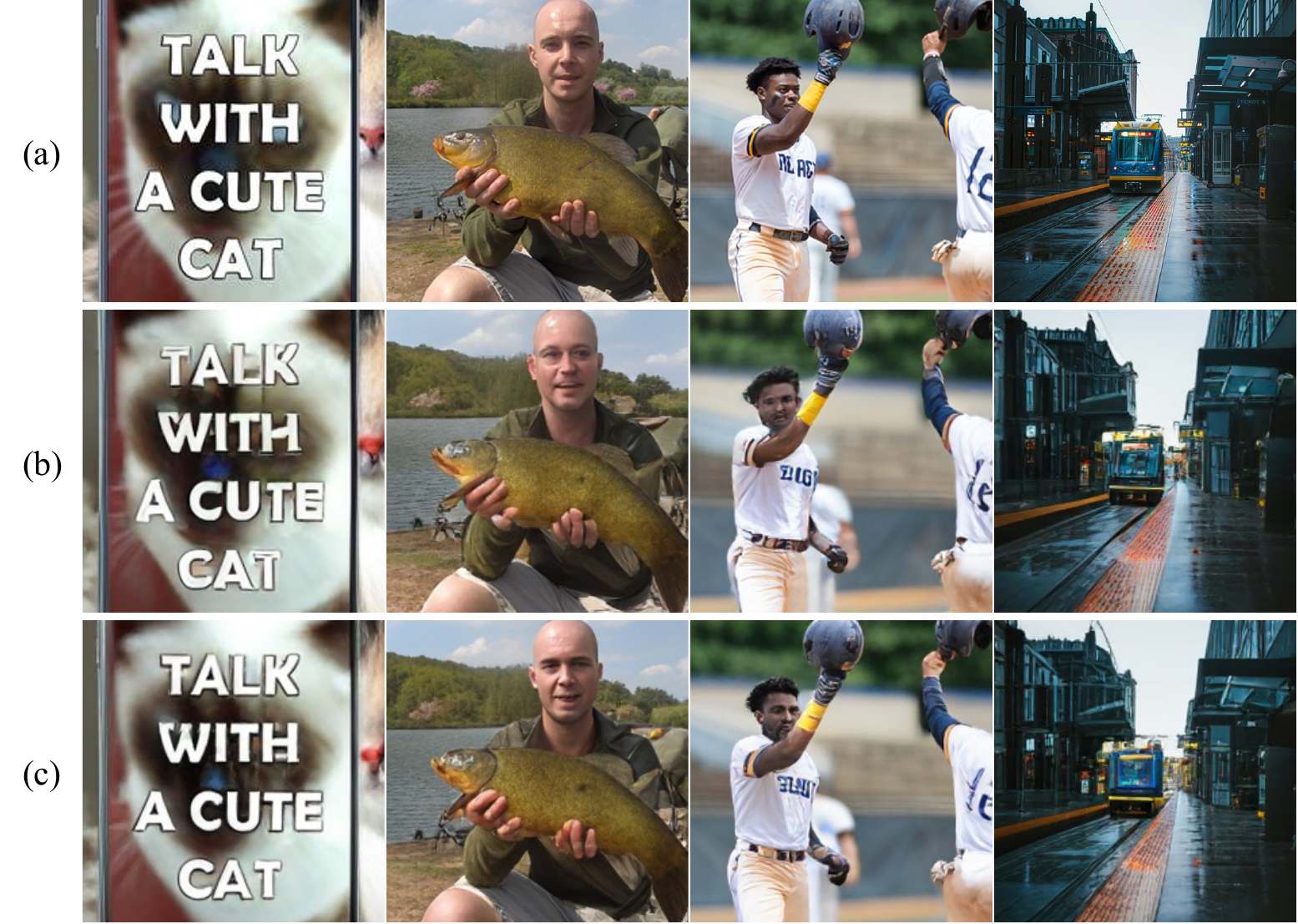}
\caption{Comparison of image reconstruction quality. (a) Original images. (b) Reconstructions using the base pixel decoder. (c) Reconstructions using the enhanced ($2\times$ capacity) decoder. The enhanced decoder demonstrates superior preservation of fine-grained details, particularly in facial details and textual elements.}
\label{fig:decoder_finetuning}
\end{figure}

\textbf{Effect of Tokenizer Decoder Finetuning.}
To further improve our model’s ability to generate fine details, we follow \cite{chang2023muse} and double both the number of residual layers and channel dimensions in the decoder.
We exclusively finetune these enhanced decoder layers while keeping all other components frozen, thereby preserving the learned visual token mappings. This enables us to improve reconstruction fidelity without compromising perception ability of TokenFlow. 
As shown in \cref{fig:decoder_finetuning}, the enhanced decoder yields notable improvements in reconstruction quality. It demonstrates superior preservation of high-frequency details, particularly in facial details and text elements.

\subsection{More Analysis of TokenFlow}
\label{appendix:additional_analysis_tokenflow}

\textbf{Analysis of Joint Distribution Learning.} 
To evaluate the effectiveness of our shared mapping mechanism, we conduct comparative experiments against VQKD \cite{peng2022beit} and VQGAN \cite{vqgan}. All models are configured with identical codebook sizes of 8,192 tokens for fair comparison. For baseline models, we utilize the official pretrained checkpoints from \cite{peng2022beit} and \cite{team2024chameleon}, respectively.
Our TokenFlow model is trained on ImageNet-1K for 50 epochs. We deliberately excludes the multi-scale VQ design \cite{var} to isolate the effects of the shared mapping in this experiment. 

\begin{figure*}[htp]
    \begin{center}
    \includegraphics[width=1.0\linewidth]{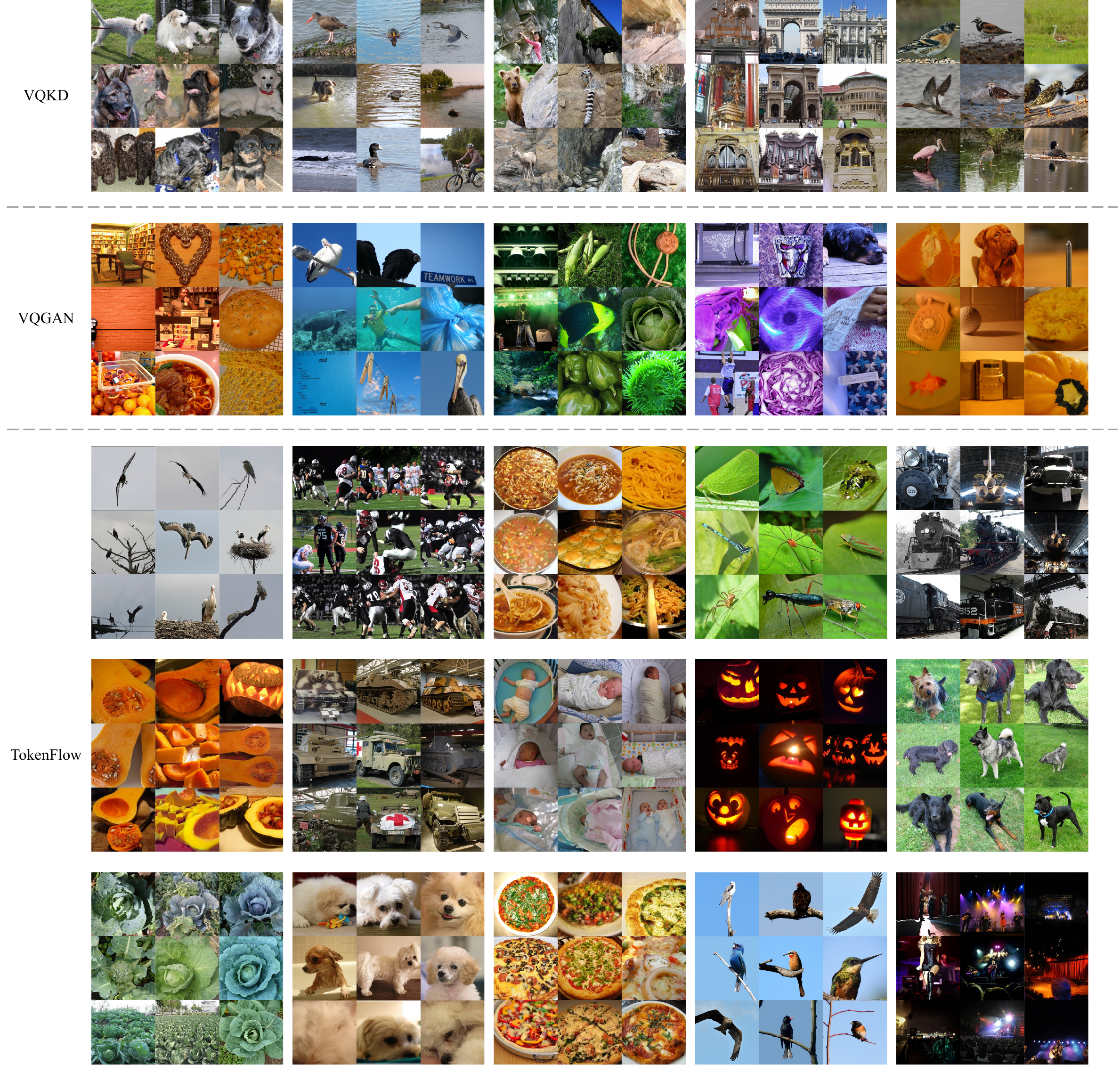}
    \caption{Qualitative comparison of images clustered by VQKD \cite{peng2022beit}, VQGAN \cite{vqgan} and our TokenFlow. VQKD clusters exhibit semantic similarity, while VQGAN clusters exhibit low-level similarity (\textit{i.e.} color and texture). Our TokenFlow can successfully combine both semantic and low-level similarity (\textit{e.g.} birds with different background can be mapped into two different index).}
    \label{fig:cluster3}
    \end{center}
\end{figure*}

For evaluation, we process 50,000 images from the ImageNet-1K validation set through each model's encoder. We apply average pooling to the extracted features to obtain a $1 \times 1$ representation, and then identify the closest index in their respective codebooks using $l_2$ distance. As shown in 
\cref{fig:cluster_distribution}, TokenFlow exhibits significantly smoother distribution against compared to others. The total non-empty clusters of TokenFlow are 7161/8192 (87.4\%), which is significantly larger than that of VQGAN (2.5\%) and VQKD (27.1\%). 
These results demonstrate that our shared mapping design enables effective learning of joint distributions across high-level semantic and low-level pixel representations. By simultaneously encoding multiple levels of visual information, we induces a joint representation space compared to single-representation architectures. This directly contributes to the superior codebook utilization observed in our experiments. Even when expanding the codebook to over 131K entries, TokenFlow maintains an exceptional utilization ratio exceeding 95\%. The clustered results is shown in \cref{fig:cluster3}.

\textbf{Automatic Balancing between Semantic Distance and Pixel Distance.} In our structure, the optimal quantize index is determined by $\argmin_i (d_{\text{sem},i} + w_\text{dis} \cdot d_{\text{pix},i})$. There exists an automatic balancing mechanism between semantic distance and pixel distance. For instance, when encountering a case where $d_{\text{sem},i}$ is relatively small while $d_{\text{pix},i}$ is large, during backpropagation, both commit loss and perceptual loss will contribute to reducing the distance between the encoded features and their quantized counterparts. This mechanism naturally narrows the gap between these two distance metrics. Therefore, we set $w_\text{dis}$ to $1.0$ across all experiments.

\begin{table*}[h]
    \centering
    \caption{Quantitative comparison of multimodal understanding capabilities between our discrete TokenFlow and their corresponding continuous semantic teachers. All experiments are trained with LLaVA-1.5 data for fair comparison. When calculating average, we use MME-P and divide it by 20 to have the same scale with other benchmarks.
    }

    \label{tab:appendix_und_results}

    \definecolor{lightblue}{RGB}{240,248,255}
    \newcommand{\colorrow}[1]{\rowcolor{lightblue} #1}
    
    \resizebox{\linewidth}{!}{
        \begin{tabular}{c | ccc | ccc ccc ccc ccc | c}
            \toprule
            Method & \# Params & Visual Encoder & Res.
            & SEEDB & MMV & POPE
            & VQAv2 & GQA & TQA 
            & AI2D & RWQA & MMMU & MMB
            & MME & MME-P & Avg.
            \\
            \midrule
            \multicolumn{17}{l}{\textit{Continuous Visual Input}}\\
            \midrule
            \multirow{3}{*}{LLaVA-1.5}
            & \multirow{3}{*}{Vicuna-13B} & CLIP ViT-B/14 \cite{clip} & 224
            & 64.1 & 30.8 & 85.1	
            & 73.8 & 61.3 & 53.4 
            & 57.8 & 50.9 & 35.1 & 62.0
            & 1737.0 & 1460.9 & 58.9
            \\
            & & ViTamin-XL \cite{chen2024vitamin} & 256
            & 65.7 & 34.6 & 85.8	
            & 76.8 & 62.6 & 57.4 
            & 59.4 & 54.4 & 35.0 & 66.4
            & 1839.1 & 1514.5 & 61.3
            \\
            & & SigLIP-SO400M \cite{siglip} & 384
            & 67.5 & 38.1 & 86.5	
            & 78.6 & 63.8 & 62.2 
            & 59.5 & 57.4 & 35.4 & 68.3
            & 1802.1 & 1488.2 & 62.9
            \\
            \midrule
            \multicolumn{17}{l}{\textit{Discrete Visual Input}}
            \\
            \midrule
            \multirow{3}{*}{Ours}
            & \multirow{3}{*}{Vicuna-13B}   & TokenFlow-B & 224
            & 60.4  & 22.4  & 84.0
            & 70.2  & 59.3  & 49.8
            & 54.2  & 49.4  & 34.2  & 55.3 
            & 1660.4  & 1353.6  & 55.2 (93.7\%)
            \\
            &   & TokenFlow-L & 256
            & 62.6  & 27.7  & 85.0
            & 73.9  & 60.3  & 54.1
            & 56.6  & 49.2  & 34.4  & 60.3
            & 1622.9    & 1365.4 & 57.5 (93.8\%)
            \\
            &   & TokenFlow-XL & 384 
            & 65.3 & 41.2 & 86.2	
            & 76.6 & 63.0 & 57.5 
            & 56.8 & 53.3 & 34.7 & 62.7
            & 1794.4 & 1502.3 & 61.1 (97.1\%)
            \\
            \bottomrule
            \end{tabular}
    }
\end{table*}

\begin{table*}[ht]
    \centering
    \caption{Comparison of generation quality on GenEval and DPG-Bench. Obj.: Object. Attri.: Attribute. $\dagger$ result is with rewriting.}
    \definecolor{lightblue}{RGB}{240,248,255}
    \definecolor{lightgray}{RGB}{229,229,229}
    \definecolor{customgray}{RGB}{200,200,200}

    \newcommand{\bluerow}[1]{\rowcolor{lightblue} #1}
    \newcommand{\grayrow}[1]{\rowcolor{lightgray} #1}
    \resizebox{\linewidth}{!}{
        \begin{tabular}{l|ccccccc|cccccc}
            \toprule
                    & \multicolumn{7}{c}{GenEval}  &  \multicolumn{6}{c}{DPG-Bench}   \\
            Method  & Overall  & Single Obj. &  Two Obj. & Counting  & Colors & Position & Color Attri.  &  Overall  & Global & Entity & Attribute & Relation & Other  \\
            \midrule
            \multicolumn{9}{l}{\textit{Diffusion-based}} \\
            \midrule
            SDv1.5 \cite{sd}    & 0.43 & 0.97 & 0.38 & 0.35 & 0.76 & 0.04 & 0.06 & 63.18 & 74.63 & 74.23 & 75.39 & 73.49 & 67.81  \\
            DALL-E 2 \cite{dalle2} & 0.52 & 0.94 & 0.66 & 0.49 & 0.77 & 0.10 & 0.19 & -- & -- & -- & -- & -- & --  \\
            SDv2.1 \cite{sd} & 0.50 & 0.98 & 0.51 & 0.44 & 0.85 & 0.07 & 0.17    & -- & -- &  --  & -- & -- & -- \\
            SDXL \cite{podell2023sdxl}  & 0.55 & 0.98 & 0.74 & 0.39 & 0.85 & 0.15 & 0.23    & 74.65 & 83.27 & 82.43 & 80.91 & 86.76 & 80.41 \\
            PixArt-alpha \cite{chen2023pixart} & 0.48 & 0.98 & 0.50 & 0.44 & 0.80 & 0.08 & 0.07    & 71.11 & 74.97 & 79.32 & 78.60 &  82.57 &  76.96  \\
            DALL-E 3 \cite{dalle3}  & 0.67$^\dagger$  & 0.96$^\dagger$ & 0.87$^\dagger$ & 0.47$^\dagger$ & 0.83$^\dagger$ & 0.43$^\dagger$ & 0.45$^\dagger$    & 83.50 & 90.97 & 89.61 & 88.39 & 90.58 & 89.83  \\ 

            \midrule
            \multicolumn{9}{l}{\textit{Autoregressive meets diffusion}} \\
            \midrule
            Show-o \cite{showo}                & 0.53 & 0.95 & 0.52 & 0.49 &  0.82 & 0.11 & 0.28 & 67.27 & 79.33 & 75.44 & 78.02 & 84.45 & 60.80  \\
            Transfusion \cite{zhou2024transfusion} & 0.63 & -- & -- & -- & -- & -- & -- & -- & -- & -- & -- & -- & -- \\

            \midrule
            \multicolumn{9}{l}{\textit{Autoregressive-based}} \\
            \midrule
            Chameleon \cite{team2024chameleon}  & 0.39 & -- & -- & --  & -- & -- & -- & -- & -- & --  & -- & -- & -- \\
            LlamaGen \cite{llamagen} & 0.32 & 0.71 & 0.34 & 0.21  &  0.58 &  0.07 & 0.04 & 64.84 & 81.76 & 75.43 & 76.17 & 84.76 & 58.40  \\

            EMU3 \cite{wang2024emu3} & 0.54 & 0.98 & 0.71 & 0.34 & 0.81 & 0.17 & 0.21 & 80.60 & 85.21 & 86.68 & 86.84 & 90.22 & 83.15  \\
            \grayrow{VAR \cite{var} & 0.53 & 0.95 & 0.60 & 0.41 & 0.81 & 0.16 & 0.24 & 71.08 & 77.51 & 78.17 & 77.80 & 85.80 & 62.00 \\}
            \multirow{2}{*}{Ours}    & 0.55 & 0.97 & 0.66 & 0.40 & 0.84 & 0.17 & 0.26 & \multirow{2}{*}{73.38} & \multirow{2}{*}{78.72} & \multirow{2}{*}{79.22} & \multirow{2}{*}{81.29} & \multirow{2}{*}{85.22} & \multirow{2}{*}{71.20}  \\
            
            & 0.63$^\dagger$ & 0.93$^\dagger$ & 0.72$^\dagger$ & 0.45$^\dagger$ & 0.82$^\dagger$ & 0.45$^\dagger$ & 0.42$^\dagger$ & & & & & & \\
            \bottomrule
        \end{tabular}
    }
    \label{tab:full_gen}
\end{table*}

\textbf{Comparison between TokenFlow and their corresponding semantic teachers.}
Table \ref{tab:appendix_und_results} presents a fair comparison between our discrete TokenFlow variants and their corresponding semantic teachers under the LLaVA-1.5 training paradigm.
TokenFlow exhibits a relative performance gap compared to its semantic teachers due to vector quantized distillation. However, this gap diminishes as resolution increases: from 6.3\% at 224$\times$224 to 6.2\% at 256$\times$256, and finally to 2.9\% at 384$\times$384. This improvement can be attributed to the increased number of discrete tokens and additional scales supplementing the residual features at higher resolutions.

\subsection{More Visual Generation Results}

\textbf{Quantitative Results.}
In \cref{tab:full_gen}, we present the complete scores for both GenEval \cite{geneval} and DPG-Bench \cite{dpgbench}. Following DALL-E 3 \cite{dalle3}, we report our GenEval results using GPT-4V as a rewriter. For DPG-Bench, we tested the results of LlamaGen and Show-o using their released checkpoints. We compare against VAR \cite{var} by using their released tokenizer and training the visual generation model under identical settings to ensure fair comparison.

\begin{figure*}[htp]
    \begin{center}
    \includegraphics[width=1.0\linewidth]{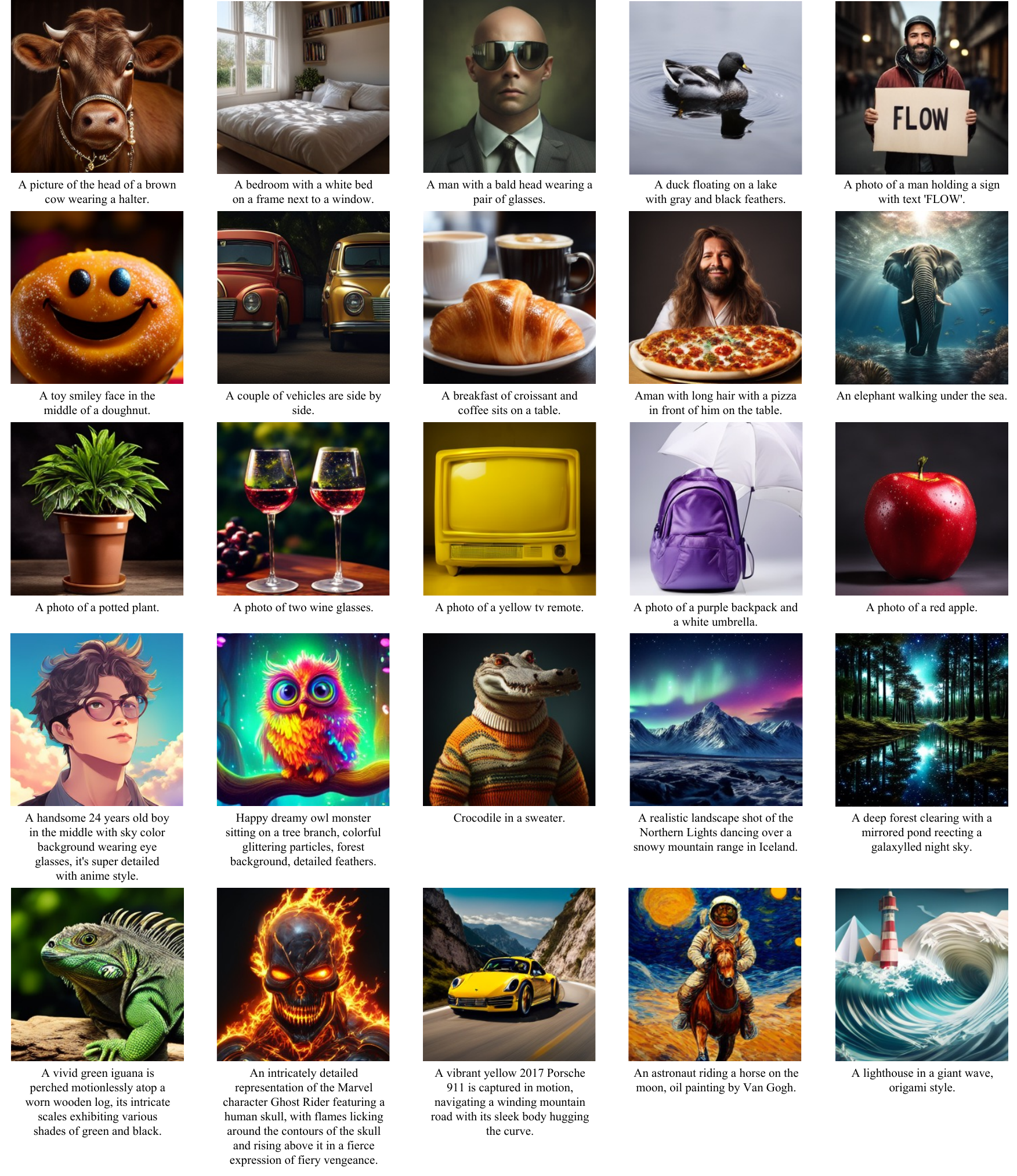}
    \caption{More Visual Generation Results with TokenFlow. We present diverse 256×256 results across various styles, subjects, and scenarios.}
    \label{fig:appendix_gen}
    \end{center}
\end{figure*}

\textbf{Qualitative Results.} We present additional visual generation results in \cref{fig:appendix_gen}. Our method can generate images with various styles, subjects, and scenarios.

\section{Limitation and Future Work}
A primary limitation of TokenFlow lies in the performance gap in multimodal understanding between our discrete tokenizer and its continuous semantic teacher, which stems from the vector quantization distillation process. While this gap narrows to 2.9\% at 384×384 resolution, several methods remain for further improvement, such as incorporating text alignment loss during tokenizer training.

In this work, we primarily focused on designing TokenFlow and validating its effectiveness separately in multimodal understanding and visual generation tasks. A natural extension of this work is the development of a fully unified model for both multimodal understanding and generation. This unification can be achieved through joint training on interleaved vision-language data. This is currently in our high
priority for exploration.

\clearpage
\clearpage

\clearpage
\clearpage

\begin{table*}[htp]
\small
\begin{center}
\caption{Detail settings of TokenFlow-B, TokenFlow-L and TokenFlow-XL.}
\label{wrap-tab:training_settings}
\resizebox{\linewidth}{!}{
\begin{tabular}{lccc}
    \toprule  
    Tokenizer & TokenFlow-B & TokenFlow-L & TokenFlow-XL \\
    \midrule
    \textit{Tokenizer settings:} &  &  & \\
    \midrule
    Input resolution & 224 & 256 & 384 \\
    Codebook size & 32,768 & 32,768 & 32,768 \\
    Semantic teacher & CLIP ViT-B/14-224 \cite{clip} & ViTamin-XL-256 \cite{chen2024vitamin} & SigLIP-SO400M-patch14-384 \cite{siglip} \\
    Multi-scale settings & [1, 2, 4, 6, 8, 10, 12, 14] & [1, 2, 3, 4, 6, 8, 10, 12, 14, 16] & [1, 2, 3, 4, 5, 6, 7, 8, 9, 10, 12, 14, 17, 22, 27] \\
    Semantic codebook embedding dimension & 32 & 32 & 32 \\
    Pixel codebook embedding dimension & 8 & 8 & 8 \\
    \midrule
    \textit{Training settings:} &  &  & \\
    \midrule
    Learning rate & 1e-4 & 1e-4 & 1e-4 \\
    Batch size & 256 & 256 & 256 \\
    Training steps & 1,000,000 & 500,000 & 500,000 \\
    Distance balance weight $w_\text{dis}$ & 1.0 & 1.0 & 1.0 \\
    Commitment loss factor $\beta$ & 0.25 & 0.25 & 0.25 \\
    Adversarial loss factor $\lambda_{\text{G}}$ & 0.5 & 0.5 & 0.5 \\
    Max gradient norm & 1.0 & 1.0 & 1.0 \\

    \bottomrule
\end{tabular}
}
\end{center}
\end{table*}

\end{document}